\documentclass{article}

\usepackage[final,nonatbib]{corl_2023} 
\usepackage[pdftex]{graphicx}
\usepackage[square,numbers]{natbib}
\usepackage{amsmath} \usepackage{amssymb}  \usepackage{amsfonts,bm}
\usepackage{siunitx}
\usepackage{xspace}
\usepackage[export]{adjustbox}
\usepackage{booktabs, array}
\usepackage{xcolor}
\usepackage{soul}
\usepackage{lib} 
\usepackage{alias} 
\usepackage{enumitem} 
\usepackage{multicol,multirow} 
\usepackage{algpseudocode}
\usepackage{algorithm}
\usepackage{wrapfig}
\usepackage{cellspace}

\usepackage[labelfont=bf,size=normal,belowskip=-15pt]{caption}

\newcommand{\mytitle}{Push Past Green: Learning to Look Behind \\ Plant Foliage by Moving It}

\title{\mytitle}

\author{
  Xiaoyu~Zhang\\
University of Illinois at Urbana-Champaign \\
  \texttt{zhang401@illinois.edu} \\
\And
  Saurabh Gupta \\
University of Illinois at Urbana-Champaign \\
  \texttt{saurabhg@illinois.edu} \\
}

\begin{document}
\maketitle

\begin{abstract}
Autonomous agriculture applications (\eg, inspection, phenotyping, plucking
fruits) require manipulating the plant foliage to look behind the leaves and the
branches. Partial visibility, extreme clutter, thin structures, 
and unknown geometry and dynamics for plants make such manipulation challenging.
We tackle these challenges through data-driven methods. 
We use self-supervision to train SRPNet, a neural network that 
predicts what space is revealed on execution of a candidate action on a given
plant. We use SRPNet with the cross-entropy
method to predict actions that are effective at revealing space beneath plant
foliage. Furthermore, as SRPNet does not just predict how
much space is revealed but also where it is revealed, we can execute a sequence
of actions that incrementally reveal more and more space beneath the plant
foliage. We experiment with a synthetic (vines) and a real plant (\dracaena) on
a physical test-bed across 5 settings including 2 settings that test
generalization to novel plant configurations. Our experiments reveal the
effectiveness of our overall method, \name, over a competitive
hand-crafted exploration method, and the effectiveness of SRPNet over a
hand-crafted dynamics model and relevant ablations. Project website with execution videos, code, data, and models: \url{https://sites.google.com/view/pushpastgreen/}.
 \end{abstract}


\section{Introduction}
\seclabel{intro}
The ability to autonomously manipulate plants is crucial in the pursuit of
sustainable agricultural practices~\cite{capellesso2016economic,
godfray2014food, foley2011solutions, davis2016precision}. Central to autonomous plant
manipulation is the {\it plant self-occlusion problem}.  Plants self-occlude
themselves (\figref{teaser}\,(left)). 
Plant leaves and branches have to be
carefully moved aside for the simplest of agriculture problems:
plant inspection, phenotyping, precision herbicide application, or 
finding and plucking fruits. This papers
tackles this plant self-occlusion problem. We develop methods that
learn to manipulate plants so as to look beneath their external foliage. 
\figref{teaser}\,(middle and right) shows steps from a sample execution 
from our method. We believe our work will serve as a building block that 
enables many different applications that require manipulation of plants 
in unstructured settings.

Manipulating external plant foliage to reveal occluded space is hard. Sensing is difficult
because of dense foliage, thin structures and partial observability. 
Control and planning is challenging because of unknown dynamics of the plant leaves and branches, and the difficulty of building a full articulable plant model.
These sensing and control challenges motivate the need for learning.
However, use of typical learning paradigms is also not straight-forward. 
Model-free RL (\eg PPO~\cite{schulman2017proximal}) requires 
interaction data at a scale that is difficult
to collect in the real world. 
Model-based RL is more sample-efficient, but is quite challenging 
here as precisely 
predicting the next observation (or state) is hard.
Imitation learning is more promising; but for the 
exploration task we tackle, the next best action depends on 
what has already been explored. This increases the amount of 
demonstration data required to train models.
Lack of high-fidelity plant simulators preclude
simulated training.

\begin{figure}[t]
\centering
\includegraphics[width=\linewidth]{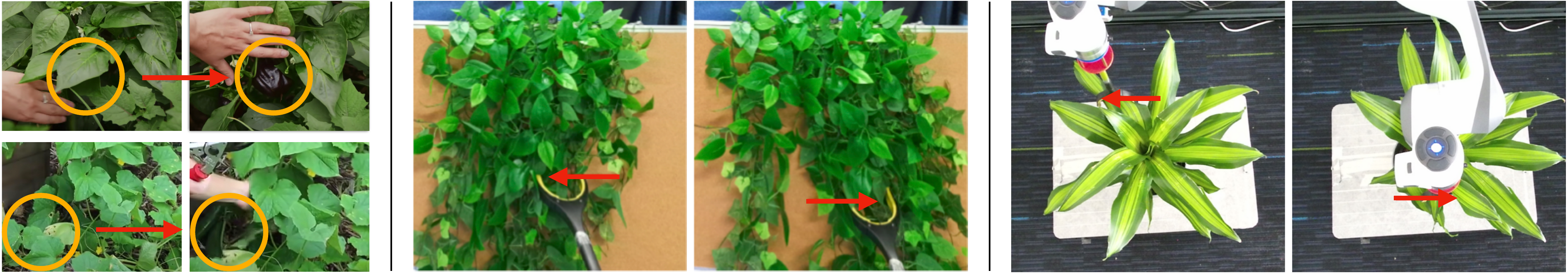}
\caption{{\bf (left)} Plants self-occlude themselves. Two examples of leaves and branches being pushed aside for inspection and picking fruits. 
This paper develops learning algorithms that enable robots to tackle this plant self-occlusion problem. 
We show actions executed by the robot to expose the space behind vines {\bf (middle)}  and \dracaena plant {\bf (right)}.}
\figlabel{teaser}
\end{figure}

Our proposal is to tackle this problem through
self-supervision~\cite{pinto2016supersizing, levine2018learning}. 
We collect a dataset of action outcomes (amount of space revealed) by letting 
the robot randomly interact with plants.  
We use this data to train a model to predict space revealed by an input action.
However, in order to derive a long-term strategy for exploring all of the space beneath the plant, the model has to predict not only how much space would get revealed, but also {\it where} (\figref{overview}\,(b), \secref{forward-model}).
In this way, the model output lets us reason about what {\it additional}
space each action would reveal. This allowing us to execute multi-step action sequences 
that explore all of the area behind the plant using a simple greedy control loop
implemented via the cross-entropy method (CEM) (\figref{overview}\,(a), \secref{control}).

This paper implements and tests these ideas on a physical platform that is
tasked with revealing space behind decorative vines and a real \dracaena plant. 
We collect 48 hours of plant interaction data and use it to train a 
neural network that we call {\it Space-Revealed Prediction Network (SRPNet)}. 
SRPNet, when used with CEM, leads to effective control strategies to reveal
all (or user-specified) space beneath the plant foliage. 
We call our overall framework {\it PushPastGreen} (PPG).

Experiments show that SRPNet outperforms a hand-crafted dynamics model and ablated versions of SRPNet.
In physical experiments, PPG outperforms a hand-crafted exploration 
strategy and versions of PPG that replace SRPNet with alternative 
choices for modeling space revealed. 
In all 5 settings across vines and \dracaena, including 2 that explicitly test
for generalization, we observe relative improvements ranging 
from 4\% to 34\% over the next best method. 
This establishes the benefits of \name and the use of learning to manipulate plants.

\section{Related Work}
\seclabel{related}

\noindent \textbf{Autonomous Agriculture.} Motivated by the need for adopting
sustainable agricultural practices~\cite{godfray2014food, foley2011solutions,
capellesso2016economic}, researchers have sought to introduce and expand the
use of autonomy for agricultural tasks~\cite{roldan2018robots, bac2014harvesting}. 
While a full review
is beyond our scope, major trends include a) development of
specialized robotic hardware~\cite{uppalapati2020berry,
mcallister2019agbots, silwal2021bumblebee}, b) development of 
algorithms for perception in cluttered agricultural
settings~\cite{freeman2022autonomous, silwal2021robust, yandun2020visual}, c)
design of control algorithms for navigation~\cite{sivakumar2021learned,
velasquez2021multi} and manipulation~\cite{xiong2020autonomous}, and d)
full autonomous farming systems~\cite{strisciuglio2018trimbot2020, xiong2020autonomous,
presten2021alphagarden, parsa2023autonomous}.

\noindent \textbf{Plant Manipulation.} For manipulation oriented tasks (\eg
fruit picking): \cite{kang2020visual} compute 3D grasp pose for largely unoccluded fruits, \cite{lehnert20183d} design a visual servoing approach to get partially occluded fruits into full view, \cite{luo2018collision, schuetz2015evaluation, tafuro2022dpmp} output trajectories for reaching fruits while avoiding collisions with plant leaves and branches, and \cite{uppalapati2020berry, soft-gripper-review} develop soft arms / end-effectors  that can maneuver around plant structures. 
Much less research actually interacts with the plant structure to accomplish tasks.
\cite{xiong2020autonomous} hand-design strategies for pushing fruits out of
the way. 
\cite{mghames2020interactive} show simulated results using
probabilistic motion primitives for pushing fruits out of the way. 
We instead study the task of looking behind plant foliage, and hand-crafted strategies proposed in~\cite{mghames2020interactive, xiong2020autonomous} are not directly applicable to our setting.
\cite{nemlekar2021robotic, bhattacharjee2014robotic} tackle reaching in plants while treating leaves as permeable obstacles, while~\cite{killpack2016model} develops efficient MPC to minimize contact forces when interacting with plants.
\cite{yao2023estimating} learn to model object's resistance to movement 
by estimating stiffness distribution from torque measurements. 
We instead directly model the effect of actions executed on the plant.

\begin{figure}[t]
\includegraphics[width=1.0\linewidth]{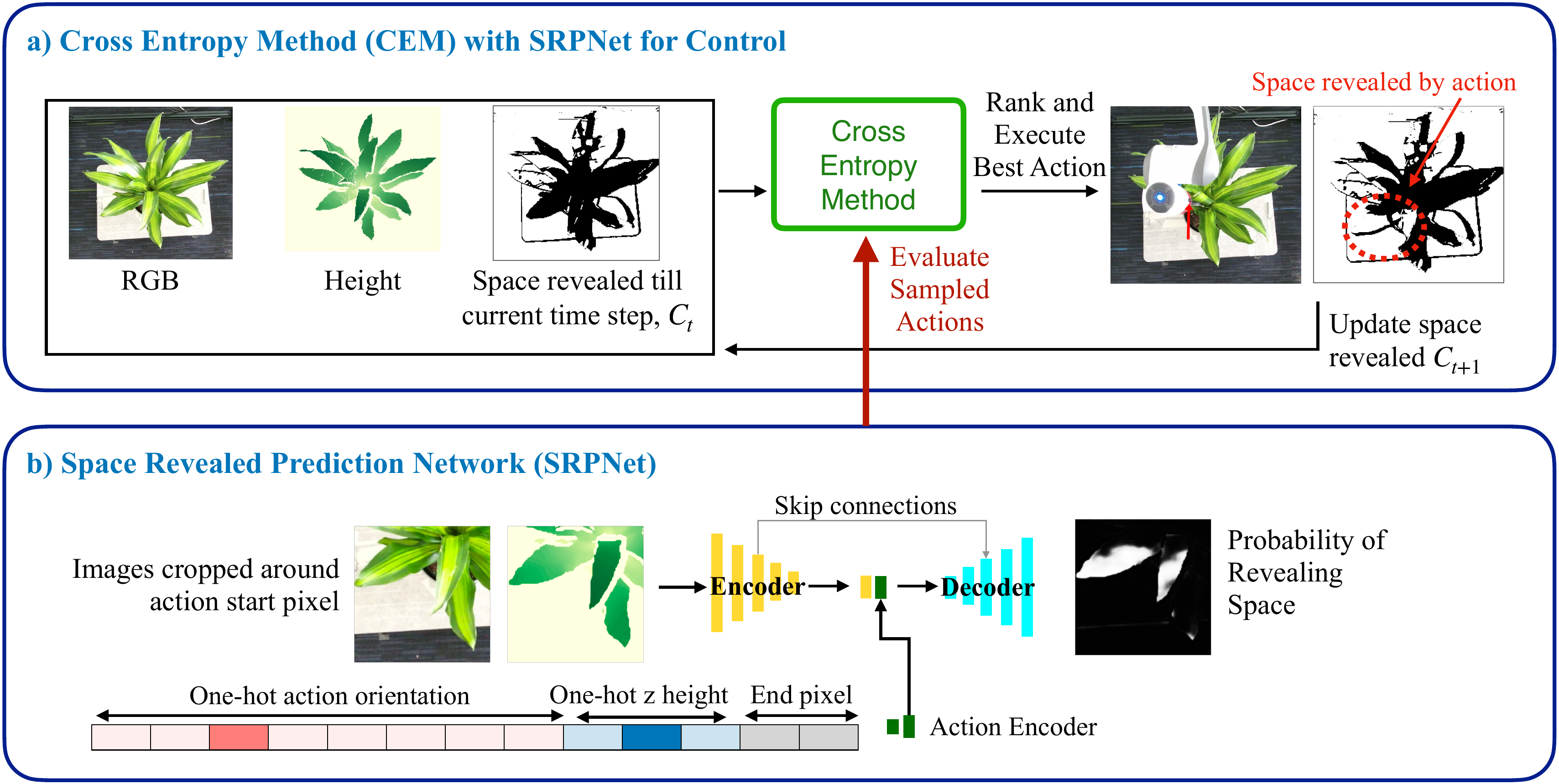}
\caption{{\bf Overview of \system.} \system learns to manipulate plants to reveal
the space behind them thus tackling the plant self-occlusion problem. \system 
includes Space-Revealed Prediction Network (SRPNet) that predicts where space is revealed upon execution of a pushing action, as shown in {\bf (b)} and described in \Secref{forward-model}. \model can not only rank actions based on how much space they will reveal, but because it can also predict
{\it where} space gets revealed, it can also be used for executing multi-step
trajectories that explore all the space behind the vines as shown in {\bf (a)} and described in \Secref{control}. \model is trained using self-supervision as described in \Secref{data-collection}.}
\figlabel{overview}
\end{figure}

\noindent \textbf{Manipulation of Deformable Objects.} Past works have
considered manipulation of other deformable objects such as
cloth~\cite{ha2022flingbot, xu2022dextairity, huang2022mesh,
weng2022fabricflownet, lin2021softgym}, ropes~\cite{chi2022iterative,
nair2017combining}, elasto-plastics~\cite{shi2022robocraft, lin2022diffskill},
fluids~\cite{lin2021softgym,li2018learning, do2018learning}, and granular
media~\cite{schenck2017learning}.  \cite{ha2022flingbot, xu2022dextairity,
chi2022iterative} design dynamic primitive actions to tackle cloth and rope
manipulation.  \cite{li2019propagation, li2018learning, shi2022robocraft, huang2022mesh} learn particle-based
forward models for deformable material and use model-based RL for control.
\cite{ lin2022planning, lin2022diffskill} compose skills for deformable object manipulation to solve long-horizon tasks. 
Our study explores plant manipulation.
Lack of high-fidelity plant simulators limits the applicability of past methods that rely on large amount of data in simulation \cite{li2018learning, weng2022fabricflownet, chi2022iterative}. At the same time, building dynamics models \cite{shi2022robocraft, huang2022mesh} for plants is hard due to dense foliage, thin branch structure, and unknown heterogeneous dynamics.

\noindent \textbf{Self-supervised Learning in Robotics.} We 
adopt a self-supervised approach for training our models.
Self-supervision techniques typically predict scalar quantities (\eg grasp outcomes~\cite{pinto2016supersizing, levine2018learning}, delta cloth coverage on workspace~\cite{ha2022flingbot}, pushing+grasping success~\cite{zeng2018learning}, \etc). 
Past work has also used self-supervision to build forward models for model-predictive control~\cite{jordan1992forward, ebert2018visual, finn2016unsupervised, agrawal2016learning} in pixel or feature spaces. 
Our work finds a middle ground. We predict not just how much space is revealed (insufficient for executing a sequence of actions), but also where it is revealed. This lets us execute sequences of actions that incrementally expose more and more space.

\section{Problem Setup}
\seclabel{problem}

\figref{problem-setup} shows the 2 different plants that we tackle, a) decorative vines vertically hanging across a board, and b) a real \dracaena plant. The vines involve a 2D exploration problem and present challenges due to
entanglement, thin structures, and extensive clutter. The real \dracaena plant
exhibits a large variation in scene depth leading to a 3D problem. The
\dracaena plant has big leaves that bend only in specific ways. Thus it requires
careful action selection. Both test cases exhibit unknown and heterogeneous 
dynamics which makes it hard to manipulate them.

As one can notice in \figsref{teaser}{problem-setup}, vines occlude
the surface behind the vines. Similarly, the \dracaena leaves
occlude the plant.  We refer to this occlusion as the {\it plant self-occlusion
problem}. The task is to have manipulation policies that can use the
non-prehensile pushing actions (as described below) to reveal the space 
beneath the plant surface.

We use the Franka Emika robot and change the end effector to a grabber (as also
done in past work~\cite{song2020grasping, young2020visual}).  We use \rgbd
cameras pointed at the plant for sensing. Our action space consists of
non-prehensile planar pushing actions (also used in past work \eg~\cite{agrawal2016learning}). We sample a 3D location and push 
in a plane parallel to the board for the vines and to the ground
for the \dracaena plant. As vines have limited depth variation, we use a fixed
$z$ for the vines, but actions 
are sampled at varying $z$ for the \dracaena plant.
\secsref{vines-action-space}{dracaena-action-space} provide more experimental details.

\section{Proposed Approach: Push Past Green}
\seclabel{approach}
\name adopts a greedy approach.
We keep track of space that has not yet been revealed, and execute actions that
would reveal the most {\it new} space. Doing this requires a model that predicts
what space a candidate action would reveal. As plants are complex to model, such
a model is hard to hand-craft. Furthermore, it is difficult to estimate the
precise state and physical parameters for plants from a single \rgbd image (\eg
placement of leaves and branches with respect to one another, location and connectivity of
all the leaves with the stems, stiffness parameters). This precludes the use of
physical simulation for such prediction. 
Thus, we design Space-Revealed Prediction Network (SRPNet), that uses learning
to directly predict space revealed on execution of a given action on a given
plant configuration (\secref{forward-model}).
Learning to directly make this prediction sidesteps the complexity of precise state estimation and physical simulation necessary to build a full dynamics model for the plant.
To obtain the data to train SRPNet, we adopt a self-supervised approach and execute random actions from the robot action space. We automatically compute the space revealed after an action
using the \rgbd image
(\secref{data-collection}).
Together with SRPNet, we design \name, a greedy algorithm that uses the
cross-entropy method (CEM)~\cite{de2005tutorial} to sample the action that promises to reveal
the most {\it new} space on top of space already revealed (\secref{control}).

\begin{figure}[t]
\centering
\includegraphics[width=\linewidth]{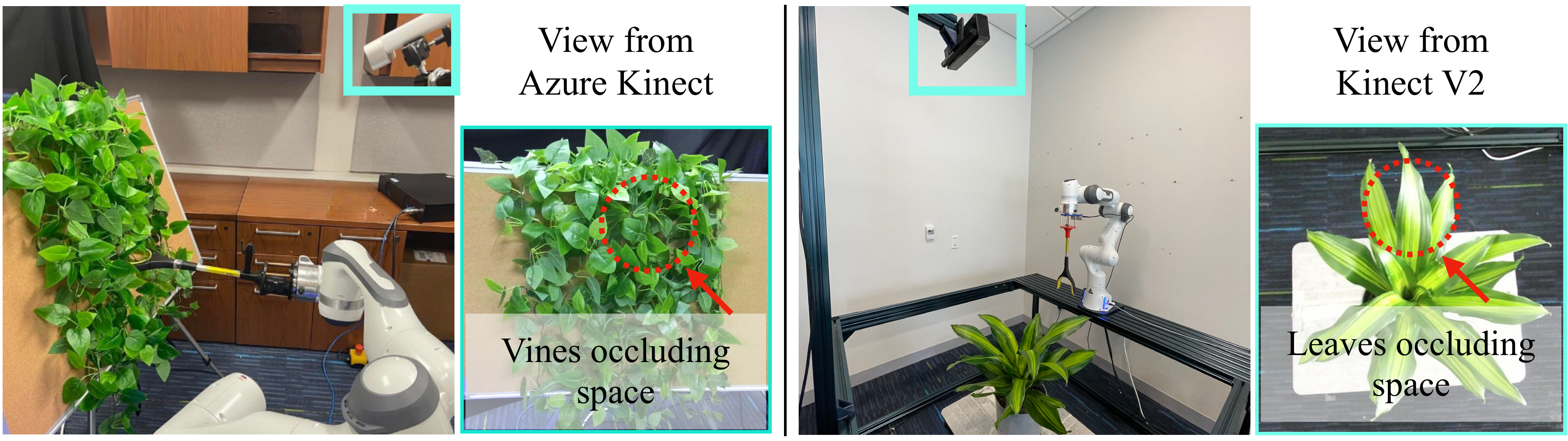}
\caption{{\bf Hardware setup for vines (left) and real \dracaena plant (right).} We use a grabber as the end-effector~\cite{song2020grasping, young2020visual}. View from the \rgbd camera is in the inset. The task is to move the vines and the \dracaena leaves aside to reveal the space occluded by them.}
\figlabel{problem-setup}
\end{figure}

\subsection{Space-Revealed Prediction Network}
\seclabel{forward-model}

\textbf{Input Representation.} As shown in \figref{overview}\,(b), the input to our model is a $200 \times 200$ patch cropped out at the action start location. 
We crop both the \rgb image and an image denoting the height relative to the surface beneath the vines (or relative to the ground beneath the \dracaena). The height image is computed using the point cloud from the \rgbd cameras.
As the model sees crops around the site of interaction, each action starts at the center of the image. We only need to represent the $z$ coordinate of the push start location, the push direction ($\theta$), and the push distance ($d$). We represent these using a) a one-hot vector depicting the push direction, b) a one-hot $z$ height, and c) push distance via the location of the action end-point \ie $[d\cos(\theta), d\sin(\theta)]$.

\textbf{Output.} The model produces an output that is the same size as the input. Each value in this spatial map represents the probability that space will get revealed at that location in the image upon execution of input action on the input plant configuration.

\textbf{Model Architecture and Loss.} 
We adopt the UNet structure~\cite{ronneberger2015u} used in image segmentation. The encoder has 5 convolution layers. The action features are processed through 2 transposed convolution layers before being concatenated with the visual features and passed to the decoder with 5 transposed convolution layers. We add a skip connection between each corresponding convolutional and transposed convolution layer. SRPNet is trained using cross-entropy loss.

\subsection{Data Collection and Preparation}
\seclabel{data-collection}
Our self-supervised data collection procedure executes random actions from the robot action space. We divide the robot's reachable space into a grid of $2\text{cm} \times 2\text{cm}$ cells. Action starting locations $(x,y,z)$ are sampled at the centers of these cells. We sample push directions and push by $15\text{cm}$ clipping to the feasible space as necessary.
Each interaction executes in about $30\text{s}$. We record \rgbd videos and robot end-effector pose over the entire duration of the interaction. 
We collected 3529 interactions for vines over 30 hours, and 2175 interactions for \dracaena over 18 hours.
We split the dataset into train, val, and test splits in a 8:1:1 ratio and train one model for each plant. 

We automatically compute ground truth for training the model on the collected data. This involves processing the \rgb and depth image before and after the interaction. For vines, we found a simple decision rule using the color value and change in depth to work well. For \dracaena, very often the entire plant wobbles upon interaction, which leads to erroneous estimates. Thus, we first align the point clouds before and after interaction and then look for depth increase to obtain ground truth.
More details are provided in Supplementary \secsref{vine-data}{dracaena-data}.

\subsection{Looking Behind Leaves Using SRPNet}
\seclabel{control}

\setlength{\intextsep}{0pt}\setlength{\columnsep}{5pt}\begin{wrapfigure}[11]{r}{0.55\linewidth}
\begin{minipage}{\linewidth}
\rule{\linewidth}{1pt}
\textbf{Algorithm 1}: \name: Revealing space beneath plants.
\label{algo}
\vspace{-5pt}
\rule{\linewidth}{0.5pt}
\begin{algorithmic}[1]
\small
\Require{Model $f$ that predicts space revealed after action}
\State Current revealed space, $C_0 \gets \text{space visible at start}$
\For{$t \gets 0$ to $T-1$}
\State Receive images $I_t$
    \State $a_t \gets \text{CEM}(C_t, f, I_t)$
    \State Execute action $a_t$
    \State Calculate additional space revealed $c_t$
    \State Update current revealed space: $C_{t+1} \gets C_t \cup c_t$
\EndFor
\end{algorithmic}
\rule[8pt]{\linewidth}{0.8pt}
\end{minipage}
\end{wrapfigure}

Algorithm~1 describes our control algorithm that uses the trained SRPNet to predict actions to reveal space behind vines. 
At each timestep $t$ of the trajectory, we use the cross-entropy method (CEM)~\cite{de2005tutorial} to pick out the best action to execute (line 4). 
We maintain the revealed space so far ($C_t$). $C_0$ is initialized to be the space visible before any actions (line~1). 
Action parameters are sampled from Gaussian distributions. 
For each candidate action, SRPNet predicts where space would be revealed. We determine {\it new} space revealed 
by subtracting the area that has already been revealed ($C_t$) from SRPNet's output.
Samples that are predicted to reveal the most {\it new space} are selected as elite which are used to fit a Gaussian distribution to sample actions for the next CEM iteration.
After all iterations, CEM outputs $a_t$, the action found to reveal the most new space (line 4). 
Upon executing $a_t$, we observe the space that is actually revealed and update $C_t$ (line 6 and 7). The process is repeated for the length of the trial.

\section{Experiments and Results}

We test our proposed framework through a combination of offline evaluations of SRPNet on our collected dataset (\secref{forward-model-eval}), and online
execution on our physical platform for the task of revealing space behind plants (\secref{physical-eval}). 
Our experiments evaluate 
a) the benefit of learning to predict space revealed by actions,
b) the effectiveness of SRPNet's input representation, 
and c) the quality of SRPNet's spatial predictions and selected actions for long-horizon and targeted exploration.

\subsection{Offline Evaluation of SRPNet}
\seclabel{forward-model-eval}

\renewcommand{\arraystretch}{1.1}
\begin{wrapfigure}[15]{r}{0.6\linewidth} 
\centering
\resizebox{\linewidth}{!}{
\setlength{\tabcolsep}{3pt}
\begin{tabular}{lccc}
\toprule
\bf Methods                                  & \bf Vines [All] & \bf Vines [5cm] & \bf \dracaena\\
\midrule
Full SRPNet (Our)                            & 46.3           & \bf 54.4       &  \bf 44.2 \\
\it Input Representation Ablations \\
$\quad$ No action                            & 30.2           & 43.5           & 28.4 \\
$\quad$ No height map                        & \bf 46.9       & 49.1           & 40.6 \\
$\quad$ No RGB                               & 33.4           & 46.4           & 28.7 \\
$\quad$ No RGB and no height map             & 28.4           & 35.2           & 10.5 \\
\it Data Augmentation Ablations \\
$\quad$ No left/right flips                  & 44.1           & 52.7           & 34.6 \\
$\quad$ No color jitter                      & 41.0           & 51.4           & 30.7 \\
\bottomrule
\end{tabular}}
\captionof{table}{{\bf Average precision for different models at predicting space revealed.} 
Higher is better. 
Our proposed input representation outperforms simpler alternatives and data augmentation boosts performance.}
\tablelabel{eval-forward}
\end{wrapfigure}
\renewcommand{\arraystretch}{1.0}
 We train and evaluate SRPNet on data gathered on our 
physical setup as described in \secref{data-collection}. 
We measure the average precision (AP) for the pixels labeled as revealed-space.
We train on the train split, select model checkpoints on the validation set, and report performance on the test set in \tableref{eval-forward}.
For vines, we report performance in two settings: 
{Vines [All]} \ie seeing the board (behind the vines) counts as revealed space, 
and {Vines [5cm]} \ie height decrease of 5cm counts as 
revealed space. For \dracaena, only seeing past the 
leaf (as determined by our automated processing from 
\secref{data-collection}) counts as revealed space.  

\textbf{Results.}
Experiments presented in \tableref{eval-forward} reveal the effectiveness of our method and provide insights about the underlying data. \textbf{First}, across all three settings, the full model is able to extract information from visual observations to produce higher quality output than just basing the predictions on the action information alone ({\it Full model} \vs { \it No \rgb and no height map}). This suggests that plant configuration (as depicted in the visual observations) is important in predicting the action outcome.
\textbf{Second}, use of action information leads to better predictions ({\it Full model} \vs { \it No action}). This suggests that different actions at the same site produce different outcomes, and SRPNet is able to make use of the action information to model these differences.
\textbf{Third}, looking at the performance across the three settings, both height map and \rgb information are useful for accurate predictions. 
There are no trivial solutions of the form `space gets revealed where height is high'. 
\textbf{Fourth}, as we only have a limited number of training samples, data augmentation strategies are effective. 
Supplementary \figref{vis-forward-model} visualizes the predictions from different versions of our model on samples from the test set. Note the nuances that our model is able to capture in contrast to the ablated versions.

\subsection{Online Evaluation for Looking Behind Plants Task} 
\seclabel{physical-eval}
We next measure the effectiveness of our proposed framework (\name w/ SRPNet) for the task of looking behind plant surface (as introduced in \secref{problem}). 
We measure the space revealed in units of 
$\text{cm}^2$ for vines and number of pixels for \dracaena.\footnote{The criterion to 
automatically determine revealed space occasionally fails. 
We manually inspected test runs to confirm and fix 
the output of the automated method. Note, that this manual inspection
is done during evaluation only. No method has access to such manual inspection, 
neither during training nor during execution.}

We start out by demonstrating that \name w/ SRPNet is able to
differentiate between good and bad actions. We then demonstrate our method on the task of looking behind plants over a 10 time-step horizon. Finally, we tackle the task of revealing space behind a user-specified spatial target.
We conduct experiments on both the vines and the \dracaena plant. 
As plants can't be exactly reset, all experiments test generalization to some extent.
To further test generalization, we explicitly evaluate performance on plant configurations that differ from those encountered during training.

Inexact resets also pose a challenge when comparing methods. 
We randomly reset the plant (such as rotating the \dracaena) between trials and expect the variance due to inexact resets to average out over multiple trials.
To prevent experimenter or other unknown environmental bias, we a)
randomly interleave trials for different methods, and b) reset {\it before}
revealing which method runs next.

\subsubsection{Baselines}
{\bf Tiling Baseline.} For the long-horizon exploration tasks, this hand-crafted baseline randomly samples from action candidates that are spread out across the workspace as shown in \figref{tiling}. We aid this baseline by limiting the action candidates to be horizontal pushes for the vines and tangential actions for the \dracaena. We found these actions to be more effective than actions in other orientations, see \tableref{dataset-statistics} for vines and \tableref{tagent-vs-random} for \dracaena.

{\bf \name w/ Other Dynamics Models.}
To disentangle if the improvement is coming from our learned SRPNet or simply from keeping track of space that has already been revealed ($C_t$) in \name, we swap SRPNet for other models in \name. Specifically, we compare to a hand-crafted dynamics model (described below) and the SRPNet No Image (\ie no \rgb and no height map) model from \tableref{eval-forward}.

We construct \textbf{hand-crafted forward models} for vines and the \dracaena plant.
These baseline models represent vines as vertically hanging spaghetti, and the \dracaena plant as 2D radially emanating spaghetti. \figref{spaghetti-model} shows the the induced free space upon action execution for this baseline model.

\begin{figure}
\centering
\begin{minipage}{.49\textwidth}
  \includegraphics[width=1.0\linewidth, trim={0cm 1cm 0 1.5cm}, clip]{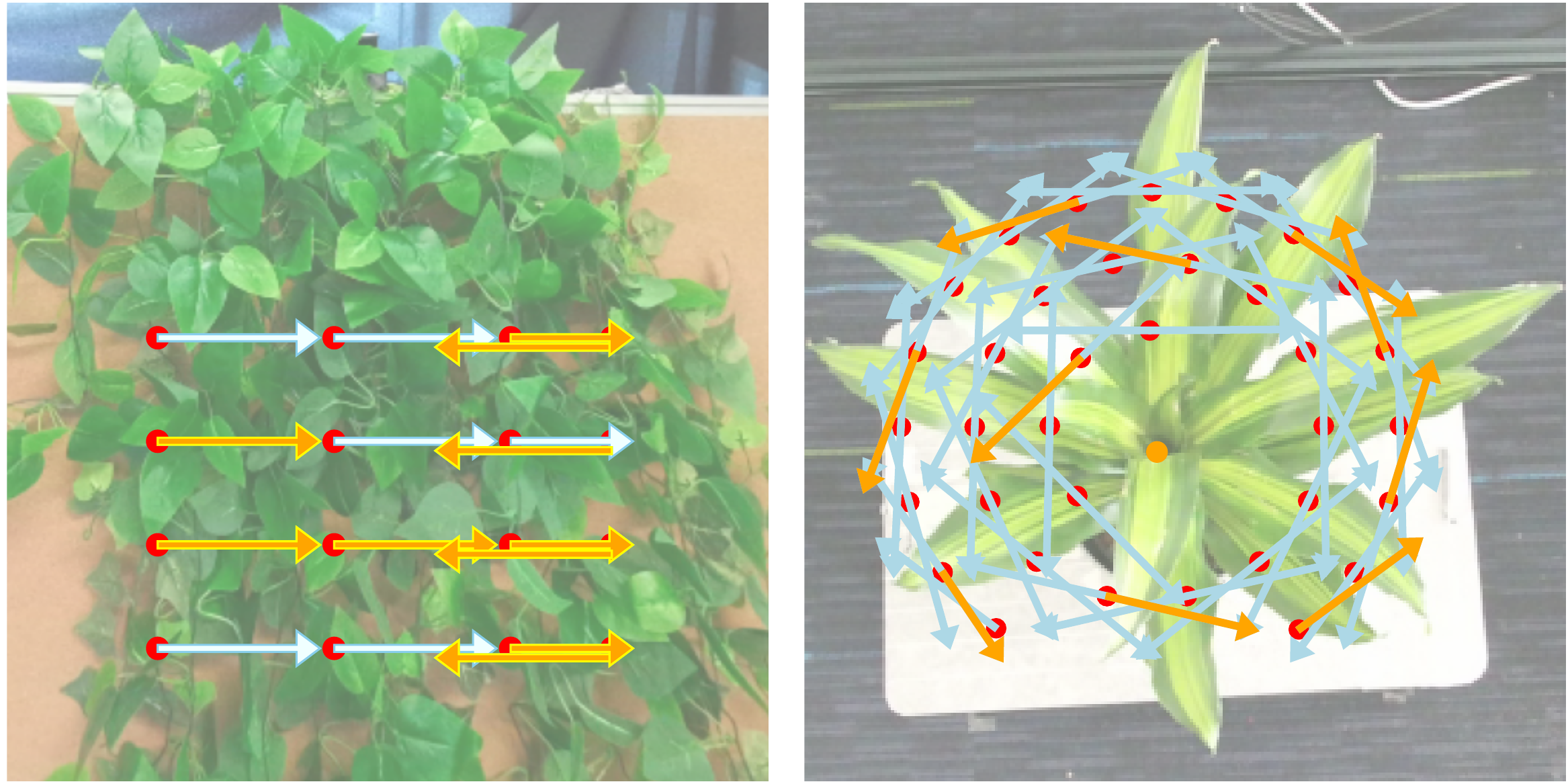}
  \captionof{figure}{{\bf Tiling baseline.} {\color{cyan} Cyan} arrows show all action candidates considered, and {\color{orange} orange} arrows show 10 actions selected during an execution.
  We aid the baseline by limiting the candidates to the most effective actions (horizontal pushes for vines, tangential pushes for Dracaena).} 
\figlabel{tiling}
\end{minipage}
\hfill
\begin{minipage}{.49\textwidth}
  \includegraphics[width=1.0\linewidth]{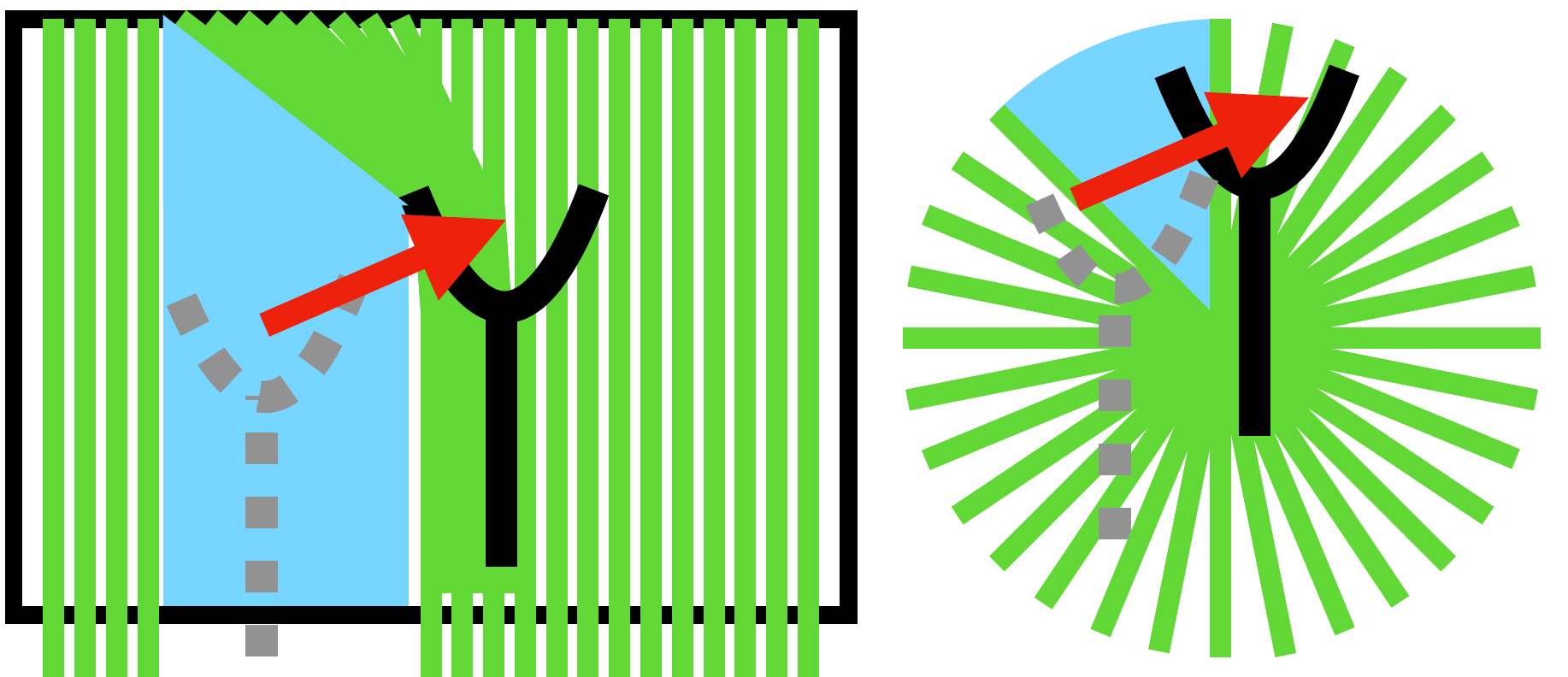}
  \captionof{figure}{{\bf Hand-crafted dynamics model} that represents vines as vertically hanging spaghetti (left), and the Dracaena plant as 2D radially emanating spaghetti (right). {\color{cyan}Cyan} area represents the space revealed by the action ({\color{red} red} arrow) under this hand-crafted model.}
  \figlabel{spaghetti-model}
\end{minipage}
\end{figure}

\subsubsection{Results}
\label{online-results}

\renewcommand{\arraystretch}{1.1}
\setlength{\tabcolsep}{3pt}
\begin{wrapfigure}[9]{r}{0.50\linewidth}
\resizebox{\linewidth}{!}{
\begin{tabular}{lcc}
\toprule
\bf Method                  & \multicolumn{2}{c}{\bf Area Revealed} \\
\cmidrule(lr){2-3}
                            & \bf Vines ($\text{cm}^2$) & \bf \dracaena (pixels)\\
\midrule
Random Horizontal /         & \multirow{2}{*}{211.7 {\scriptsize [184.2, 255.8]}} & \multirow{2}{*}{5125.6 {\scriptsize [3423.8, 7147.0]}}\\
$\quad$ Tangential Action   \\  
\name w/ SRPNet (Our)       & \bf 344.1 {\scriptsize [320.4, 372.3]}  
& { \bf 7644.4 {\scriptsize [6335.6, 9057.5]}} \\
\bottomrule
\end{tabular}}
\captionof{table}{{\name selected actions are more effective at revealing space.}}
\tablelabel{first-action}
\end{wrapfigure}
\setlength{\tabcolsep}{6pt}

 \textbf{Single Action Selection Performance.}
\tableref{first-action} compares the effectiveness of \name w/ SRPNet 
at picking an action that reveals the most occluded space, against random actions from
the robot's action space. For the strongest comparison, we limit the random sampling to the most effective actions, horizontal pushes for the vines and tangential pushes for the \dracaena, as shown in \figref{tiling}. 
\tableref{first-action} reports the average space revealed (along with 95\% confidence interval) over at least 20 trials for each method. 
Our approach leads to a relative improvement of 62\% for vines and 49\% for \dracaena over this strong baseline. This suggests that our model
is able to interpret visual observations to identify good interaction sites.

\renewcommand{\arraystretch}{1.2}
\setlength{\tabcolsep}{3pt}
\begin{wrapfigure}[16]{r}{0.55\textwidth}
\includegraphics[width=\linewidth]{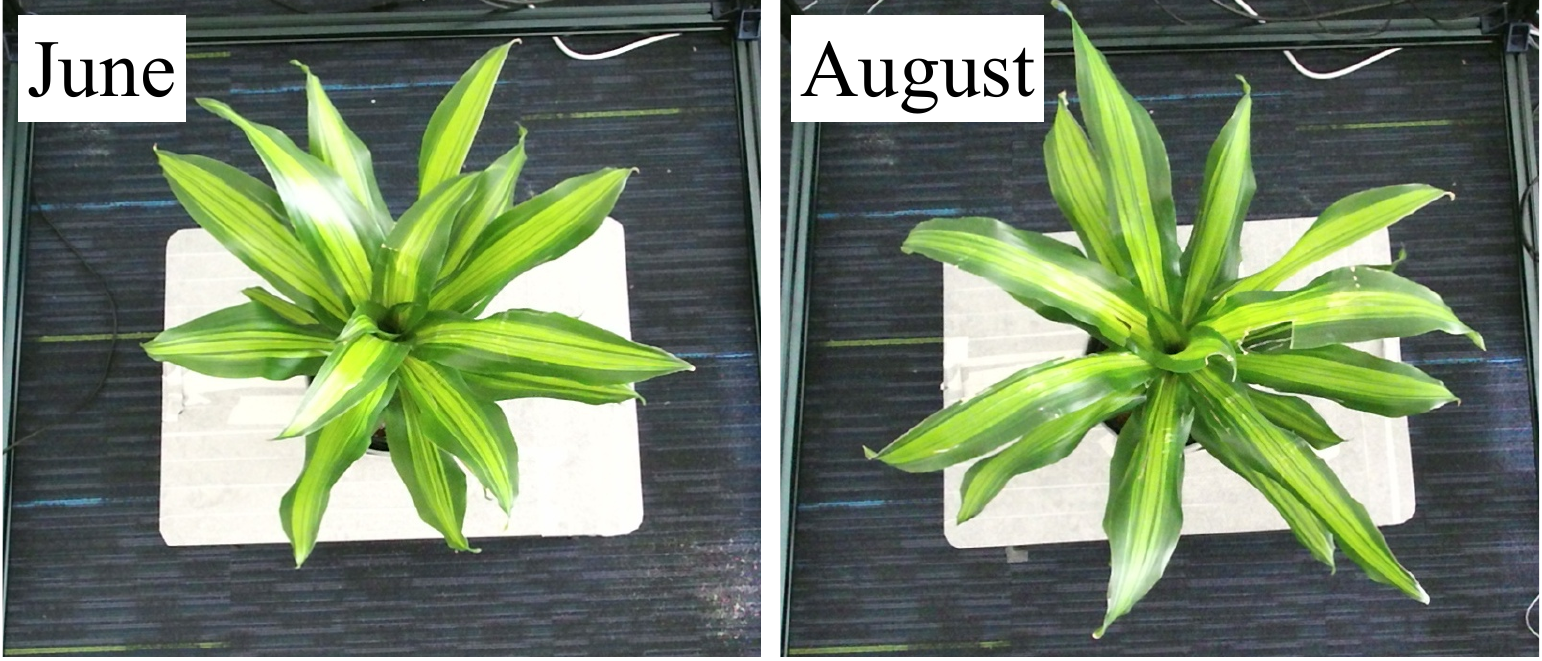} \\
\resizebox{\linewidth}{!}{
\begin{tabular}{lcc}
\toprule
\bf Method                  & \multicolumn{2}{c}{\bf Area Revealed (pixels)} \\
\cmidrule(lr){2-3}
                            &  \bf June & \bf August \\
\midrule
Random Tangential Action         & 5125.6 {\scriptsize [3423.8, 7147.0]} & 6780.7 {\scriptsize [4399.3, 8618.1]}\\
\name w/ SRPNet (Our)       & {\bf 7644.4 {\scriptsize [6335.6, 9057.5]}} & {\bf 11551.4 {\scriptsize [9464.7, 14644.9]}} \\
\bottomrule
\end{tabular}}
\caption{{We evaluate a model trained with data collected from the June Dracaena on the August Dracaena.}}
\figlabel{plant-growth}
\end{wrapfigure}
\setlength{\tabcolsep}{6pt}
 \textbf{Generalization across Plant Growth Performance.}
To test generalization, we run the single action selection experiment
(\tableref{first-action}) on the Dracaena plant after two-month of growth (in
August) but with a model trained on data from June. 
Note the difference between plants in \figref{plant-growth}\,(top). Despite changes
in appearance and leaf length, \name generalizes to the grown Dracaena and
outperforms random tangential action as shown in
\figref{plant-growth}\,(bottom).

\begin{figure}[b]
    \centering
    \small
    \setlength{\tabcolsep}{0.1em} \includegraphics[width=1.0\linewidth]{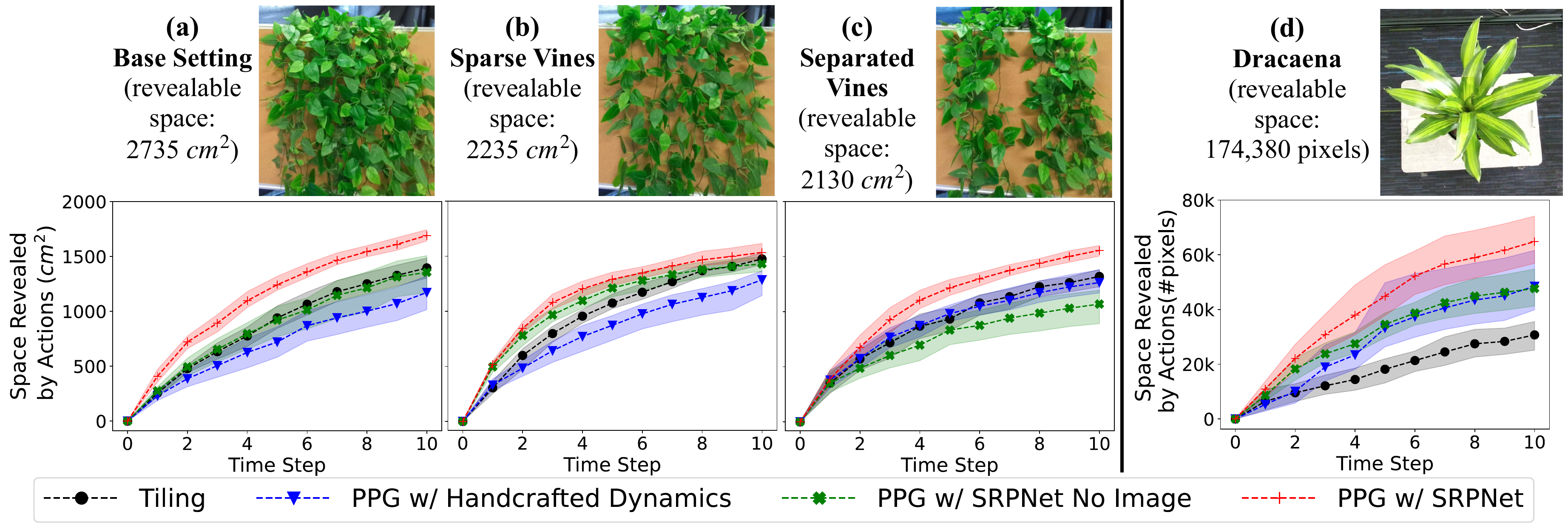}
\caption{{\bf Comparison of different methods for multi-step exploration of space behind plant foliage.} We show results in four settings across vines (left) and \dracaena (right). The line plots show average cumulative space revealed by actions up to time step $t$ 
across 10 trials (along with 95\% confidence intervals). SRPNet training data was collected in the base setting shown in (a) for vines and (d) for \dracaena. (b) and (c) are novel settings that test generalization capabilities of our model. Our method (\name w/ SRPNet) outperforms all baselines (a strong hand-crafted policy, and \name with other dynamics models) across all settings.}
\figlabel{comparison}
\end{figure}

\textbf{Long-horizon Exploration Performance.} 
Next, we study if SRPNet can be used for situations that require multiple sequential interactions to reveal space behind plants. The task is to maximize the cumulative space revealed over a 10 time-step episode. This further tests the quality of SRPNet which now also needs to accurately predict {\it where} it thinks space will be revealed. 

We conduct 4 experiments, one on \dracaena and 3 on vines.
For the vines, we considered 3 settings: a) Base Setting: vine setting as used for collecting training data, and 2 novel settings to test generalization: b) Sparse Vines, and c) Separated Vines. While the last two settings explicitly test generalization, we note that the first setting also tests models on novel vine configurations not exactly seen in training. For \dracaena, we only conducted experiments in the Base Setting. 

\figref{comparison} plots the average space revealed (in $\text{cm}^2$ for vines and in pixels for \dracaena) as a function of the number of time-steps. We report the mean over 10 trials and also show the 95\% confidence interval. Across all three experiments our proposed method achieves the strongest performance. 
Supplementary \figref{vis-execution} and videos on the website show some sample executions.

Results suggests that SRPNet is quite effective at predicting where space will get revealed (\name w/ SRPNet \vs Tiling). 
Learning and planning via CEM lets us model complex behavior which is hard to hand-craft. 
Improvements over the tiling baseline increase as the action space becomes larger (\dracaena \vs vines).
Moreover, benefits don't just come because of keeping track of revealed space ($C_t$), but also from the use of SRPNet (\name w/ SRPNet \vs \name w/ Handcrafted Dynamics).
Furthermore, our model is able to interpret the nuances depicted in the visual information to predict good actions (\name w/ SRPNet \vs \name w/ SRPNet No Image). 
SRPNet also leads to benefits in novel vine configurations. Benefits are larger in the separated vines case than for the sparse vines. This may be because the separated vines are still locally dense and SRPNet processes local patches.

\begin{wrapfigure}[10]{r}{0.5\textwidth}
\centering
\includegraphics[width=\linewidth]{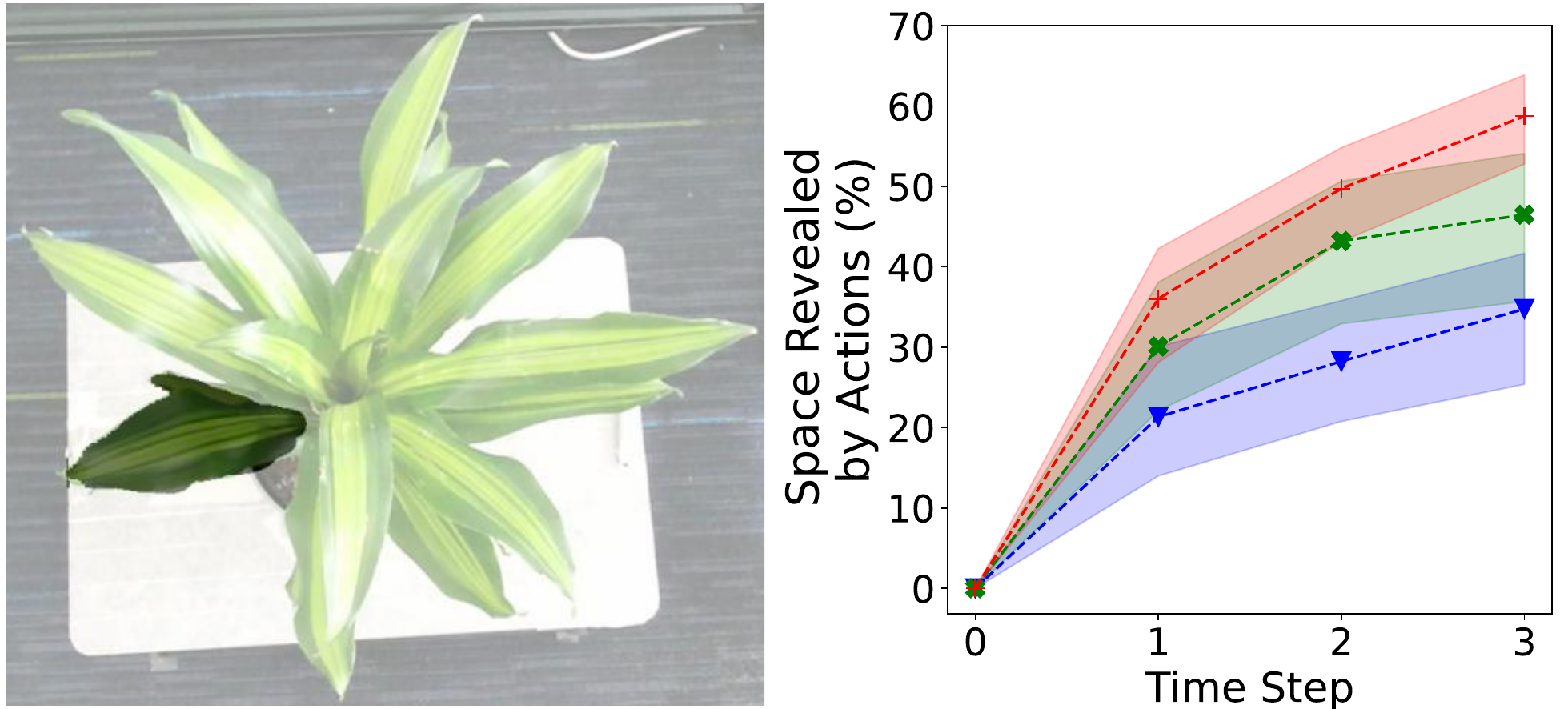}
\caption{{\name is also effective at revealing space behind a specific spatial target.}}
\figlabel{targetted}
\end{wrapfigure} \textbf{Targeted Revealing Performance.} Our final experiment tackles the task of targeted exploration. The task here is to reveal space at a user-defined region, $m$. \figref{targetted}\,(left) shows a sample user-selected region. We tackle this task by setting $C_0$ to be $\bar{m}$, the complement of the user-defined region. \figref{targetted}\,(right) presents the results (same legend as for \figref{comparison}, but without Tiling Baseline). Again, as SRPNet reliably models the effect of actions, \name with SRPNet outperforms \name with other dynamics models.

\section{Discussion} 
\seclabel{discussion}
In this paper, we introduced \name and SRPNet to tackle the
problem of manipulating the external plant foliage to look within 
the plant (the {\it plant-self occlusion problem}). SRPNet uses 
self-supervised learning to model
what space is revealed upon execution of different actions on plants.
This sidesteps the difficulty in perception arising from dense foliage, 
thin structures, and partial information. 
\name derives control strategies using SRPNet via CEM, to output
sequence of actions that can incrementally explore space occluded by
plants.
Experiments on a physical platform demonstrate the benefits of our proposed learning framework
for tackling the plant self-occlusion problem.

\section{Limitations} 
We believe ours is a unique and first-of-its-kind study, but it has its limitations.
We note two failure modes. First, 
\name sometimes resamples an overly optimistic action 
(that doesn't actually reveal much space, so nothing changes and CEM returns a very similar next action) 
many times over without making progress. 
Second, as each individual push action 
doesn't use visual feedback it can't recover from say when a leaf slips from below the gripper.
These may be mitigated by incorporating spatial diversity while selecting actions 
and by learning closed loop leaf manipulation policies through imitation. 
More generally, our overall approach relies on input from 
\rgbd cameras that are known to perform poorly in the wild. This may be mitigated
through use of specialized stereo cameras built for farm settings~\cite{georgekantorcamera}.
Our techniques for automatic estimation of revealed space can be improved further 
using recent point tracking models~\cite{harley2022particle}, and it may be useful to build models
that can predict and keep track of full 3D space.
Experiments should be conducted with more diverse real plants. 
Future work should also rank actions from the perspective
of the damage they cause to the plant, perhaps via some tactile sensing~\cite{bhattacharjee2014robotic}. Lastly, while 
autonomous agriculture provides a path towards sustainable
agricultural practices, societal impact of such automation should be studied 
before deployment.

\acknowledgments{
Images in \figref{teaser}\,(top) have been taken from
\href{https://www.youtube.com/watch?v=M9eT4DJLUB4&t=532s}{YouTube video 1} and
\href{https://www.youtube.com/watch?v=Oa2apuOj5R4&t=984s}{YouTube video 2}.
This material is based upon work supported by the USDA/NSF AIFARMS National AI
Institute (USDA \#2020-67021-32799), an NSF CAREER Award (IIS-2143873), an
Amazon Research Award, and an NVidia Academic Hardware Grant. We thank Matthew Chang and Aditya Prakash for helpful feedback. We thank Kevin Zhang for help setting up robot experiments.}

\bibliography{biblioLong, references}  

\clearpage

\def\thesection{S\arabic{section}}
\def\thetable{S\arabic{table}}
\def\thefigure{S\arabic{figure}}
\setcounter{figure}{0}
\setcounter{table}{0}

\appendix

\section*{Push Past Green: Learning to Look Behind Plant Foliage by Moving It \\ Supplementary Material}

\section{Implementation Details for Vine Experiments}

\captionsetup{belowskip=0pt}
\subsection{Robot Action Space}
\seclabel{vines-action-space}
\begin{figure}[h]
\centering
\includegraphics[width=1.0\textwidth]{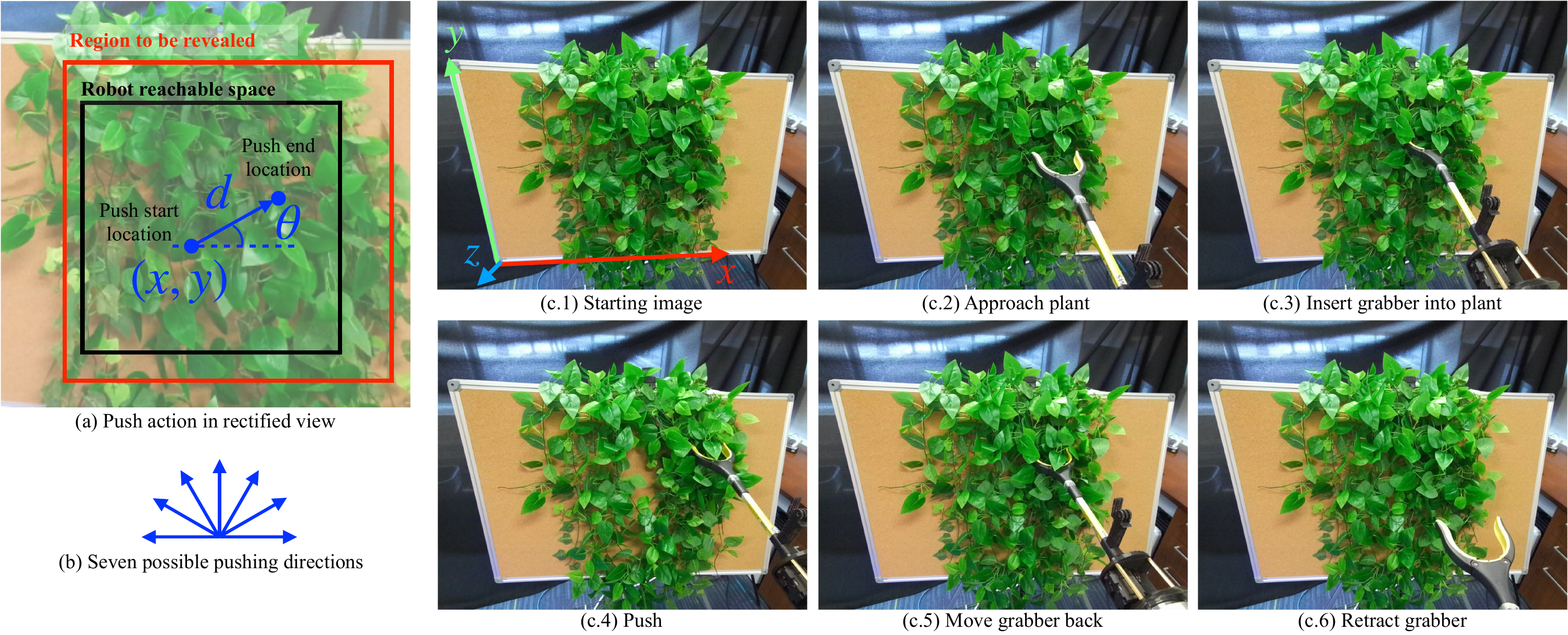}
\caption{{\bf Robot's action space for vine setup.} {\bf(a)} shows the rectified image that we operate in, the region to be revealed (red box), and the region that the robot can reach (black box). The robot can execute push actions that start at a pixel $(x,y)$ in the rectified image and push a distance of $d$ at an angle $\theta$. We use 7 discrete push directions $\{0, \pi/6, \pi/3, \pi/2, \ldots, \pi\}$ as shown in {\bf (b)}. {\bf (c.1) through (c.6)} show a sample execution of the push action.}
\figlabel{robot-action-space-vines}
\end{figure} 

\vspace{8pt}
The robot's action space consists of non-prehensile pushing actions. As shown in \figref{robot-action-space-vines}\,(a), 
these actions are parameterized by $(x, y, \theta, d)$. 
Such parameterization for pushing actions has been used in past works, \eg~\cite{agrawal2016learning}. 
Here, $(x,y)$ denotes the start location for the push interaction on the board, $\theta$ denotes the push angle, and $d$ denotes the push length. 
As shown in \figref{robot-action-space-vines}\,(b), we sample $\theta$ to be one of 7 angles from $\{0, \pi/6, \pi/3, \pi/2, 2\pi/3, 5\pi/6, \pi\}$. 
We do not sample angles greater than $\pi$ because pushing towards the bottom of the vines only drags down the vines and could pull the board over. 
We assume that the grabber inserts deep enough into the vines to push the vines but not too far to knock it over; therefore, the pushes are planar actions executed with the same $z$ value.
We estimate the location and orientation of the board and establish a coordinate frame that is aligned with the board. Push locations and orientations are expressed in this coordinate frame. 
We implement these actions by moving the grabber through 4 waypoints, as shown in \figref{robot-action-space-vines}\,(c.2) to \figref{robot-action-space-vines}\,(c.5). In \figref{robot-action-space-vines}\,(c.4), we can see the effect of a randomly sampled action on the state of the vines.
We drive the Franka Emika robot between these waypoints using the Franka-interface and frankapy library~\cite{zhang2020modular}.

\subsection{SRPNet}
For the vine setup, we are unable to position the camera such that it is perpendicular to the board.
Therefore, we design SRPNet to work on rectified images of the scene, such that the camera is looking straight at the vines. This corresponds to using a homography to transform the image such that the surface underneath the vines becomes fronto-parallel. 
We build the model to only reason about a $40\text{cm} \times 40\text{cm}$ neighborhood around the action start location. 
Parts of the board get occluded behind the robot arm as the robot executes the action. These occluded parts and area with no depth readings are masked out for evaluation and training. 

\subsection{Data Collection}
\seclabel{vine-data}

\renewcommand{\arraystretch}{1.1}
\begin{figure}[t]
\setlength{\tabcolsep}{4pt}
\centering
\resizebox{\linewidth}{!}{
\begin{tabular}{lcccccccc}
\toprule
\bf Push Angle & 0 & $\pi/6$ & $\pi/3$ & $\pi/2$ & $2\pi/3$ & $5\pi/6$ & $\pi$ & Full Dataset\\ \midrule
\bf \# Interactions        & 985   &  460  &  360  &  348  &  359   &   433  &   584  &  3529    \\
\bf Mean area revealed ($\text{cm}^2$) & 215.7  &  177.3  &  93.6  &  58.8  &  100.4  &  180.9  &   237.1  &    170.3  \\
\bottomrule
\end{tabular}}
\captionof{table}{{\bf Statistics for the different push directions in the
collected vine dataset.} Collected dataset reveals many aspects of the problem. For example, for vines, horizontal push actions ($0$ and $\pi$) are the most effective
at this task.}
\tablelabel{dataset-statistics}
\end{figure}
\renewcommand{\arraystretch}{1.0}
 
The robot's actions are in the same fronto-parallel plane used for SRPNet as described earlier. We estimate the space that can be safely reached by the robot ahead of time to make sure it is not close to its joint limits during interactions. The resulting space is roughly $40\text{cm} \times 40\text{cm}$. We divide the feasible space into a $20\times 20$ grid. Action starting locations $(x,y)$ are sampled at the centers of these grid squares (\ie, 400 possible starting locations). We sample push directions from the 7 possible angles, $\{0, \pi/6, 2\pi/6, \ldots, 6\pi/6\}$, and push by $15\text{cm}$ clipping to the feasible space as necessary. Therefore, not all interactions have $d=15$; for starting locations near the boundary, $d < 15$.

Our full dataset contains 3529 interactions (summed to roughly 30 hours) collected over 11 different days (nonconsecutive). 
This data includes 2571 interactions done specifically for the purpose of data collection. The remaining interactions come from when we were developing control algorithms. 
These don't follow uniform sampling from the robot's action space and are biased towards horizontal actions since the most effective actions for the baselines are often horizontal actions.

We automatically compute the ground truth for training the model on the collected data. Specifically, we use color thresholding to determine when the surface beneath the vines has been fully exposed. We found this simple strategy to be reasonably robust. Note that while we train and use SRPNet to predict whether {\it all} vines were moved aside to reveal the board, we can process the data in other ways to also train the model for other tasks. For example, we can re-purpose the data for a task that involves only looking beneath the first layer of vines. We can re-compute ground truth to identify locations where the height decreased by (say) more than $5\text{cm}$ for such a task. 

\subsection{Cross-entropy Method}

Our CEM implementation uses 3 iterations that each evaluate 300 candidate actions. We sample $(x,y,\theta)$ from Gaussian distributions. 
In the first CEM iteration, $x,y,\theta$ are sampled from Gaussians with different mean and variances, chosen to cover the whole action space. The parameters are then discretized to match the distribution from  data collection. 
When sampling actions, we only retain action samples that are feasible (\ie within the robot's reachable space as shown in \figref{robot-action-space-vines}\,(a)). 
Elite samples are the top $20\%$ candidates that have the most amount of \textit{new space} revealed. Running line 3 to 6 in Algorithm 1 (\secref{control}) for vines takes about 5 seconds.  

\clearpage 
\section{Implementation Details for Dracaena Experiments} 

\subsection{Robot Action Space}
\seclabel{dracaena-action-space}

The robot's action space for \dracaena is similar to that of vines.
However, since the Dracaena leaves are at different heights, we define three possible $z$ values that the grabber can insert to. 
The \dracaena plant body is about 45cm tall so we defined the $z$ values to be about 22.5, 17.5, and 12.5cm from the top of the plant.
For each $z$ value, planar pushing actions $(x,y,\theta,d)$ are defined on a plane parallel to the ground. 
We sample $\theta$ from 8 possible angles: $\{0, \pi/4, \pi/2, 3\pi/4, \pi, 5\pi/4, 3\pi/2, 7\pi/4\}$. The angles are 45 degrees away from one another instead of 30 degrees as used in vines because we want to keep the total number of possible actions reasonable.  

\begin{figure*}[h!]
\centering
\includegraphics[width=1.0\textwidth]{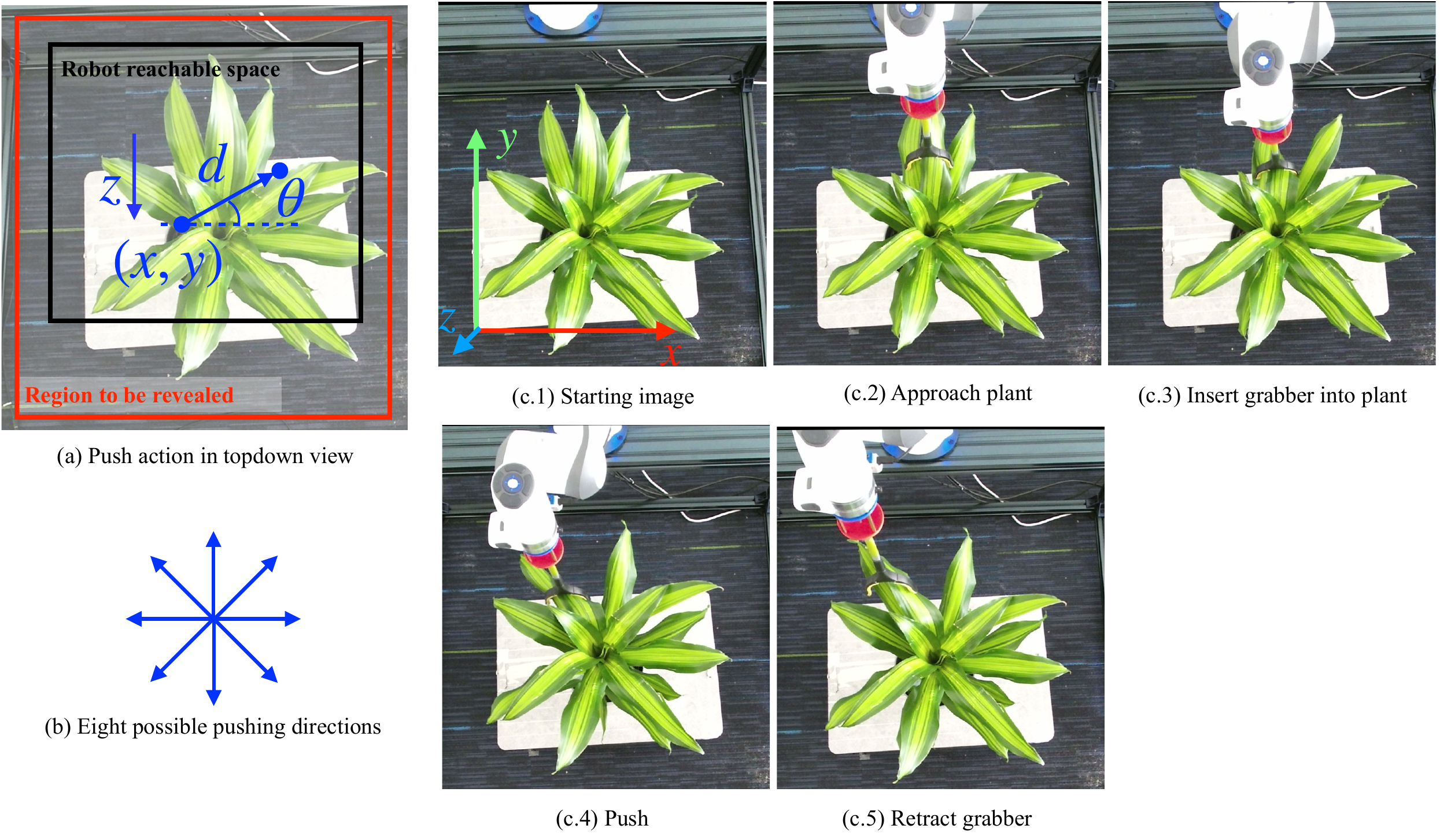}
\caption{{\bf Dracaena robot action space.} Similar to \figref{robot-action-space-vines}, {\bf(a)} shows the image from the camera, {\bf(b)} shows the pushing directions, and {\bf(c)} shows the sample execution of a push action.}
\figlabel{dracaena-robot-action-space}
\end{figure*}

\subsection{SRPNet}
Since the Kinect camera is looking down at the \dracaena plant, SRPNet does not work on rectified images as it does for vines and instead takes in images from the camera as they are. We project action start locations into their image coordinates using the camera intrinsics and crop around the locations to obtain local patches to input into the network. When training SRPNet, adding another head to predict height decrease in addition to the binary classification head helps AP performance. We use Huber loss with $\delta=0.1$ to provide an auxiliary loss to the network.    

\subsection{Data Collection}
\seclabel{dracaena-data}

The reachable space of the robot in the \dracaena setup is roughly $57\text{cm} \times 53\text{cm}$ and corresponds to a $29 \times 27$ grid of $2$cm cells. 
Similar to the vines' setup, the action starting point $(x,y)$ is sampled from these 783 possible locations. 
Given that pushing from the center of the plant tends to displace it entirely, we aim to discourage such actions to prevent damage to areas where new leaves may sprout.
We manually delineate a rectangular region around the plant center and do not sample or execute actions in this region. 
We also sample $z$ from 3 possible values (22.5, 17.5, and 12.5cm from the top of the plant as mentioned before), push directions from 8 possible angles, $\{0, \pi/4, \pi/2, 3\pi/4, \pi, 5\pi/4, 3\pi/2, 7\pi/4\}$, and push by $15\text{cm}$ clipping to the feasible space as necessary. Therefore, not all interactions have $d=15$; for starting locations near the boundary, $d < 15$.

Since the plant wobbles during pushing, we discount the area that is revealed due to whole-plant movement. We construct plant point clouds before and after an action; then, iterative closest point (ICP) is performed to align the two point clouds. 
During execution, the robot body occludes parts of the plant, so we mount a Intel RealSense camera at the wrist to fill in these occluded regions to aid ICP.
Area where the plant height has decreased in the aligned point cloud is considered to be revealed space.      

\renewcommand{\arraystretch}{1.1}
\begin{figure}[t]
\setlength{\tabcolsep}{4pt}
\centering
\resizebox{\linewidth}{!}{
\begin{tabular}{lccccccccc}
\toprule
\bf Push Angle & 0 & $\pi/4$ & $\pi/2$ & $3\pi/4$ & $\pi$ & $5\pi/4$ & $3\pi/2$ & $7\pi/4$ & Full Dataset \\ 
\midrule
\bf \# Interactions        & 257   &  262  &  295  &  273  &  249   &   297  &   289 &   253  &  2175    \\
\bf Mean area revealed (pixels) & 1391.4  &  1138.7  &  990.4  &  802.0  & 1154.0  &  1110.8  &  1154.9  &  1495.2  &  1147.7    \\
\bottomrule
\end{tabular}}
\captionof{table}{{\bf Statistics for the different push directions in the
collected \dracaena dataset.}}
\tablelabel{dracaena-dataset-statistics}
\end{figure}
\renewcommand{\arraystretch}{1.0}
 
\subsection{Cross-entropy Method}
We follow the same algorithm as the one outlined in Algorithm 1 (\secref{control}). The \dracaena CEM uses 3 iterations that each evaluate 300 candidate actions. We sample $(x,y,\theta,z)$ from uniform distributions within the robot's reachable space. The parameters are then discretized to match the data collection's distribution. Top $20\%$ candidates that reveal the most amount of new space are chosen as elite samples that are fitted with Gaussian distributions for the next iteration. Running one iteration takes about 7 seconds.     

\subsection{Comparing Tangential to Random Actions}

\renewcommand{\arraystretch}{1.1}
\begin{figure}[h!] 
\centering
\begin{tabular}{llc}
\toprule
\bf Method                  & \phz{~~~~~~~~~~~~~~~} & \bf Area revealed (pixels) \\
\midrule
Random Action    &              & 3956.1 {\scriptsize [$2826.1$, $5045.9$]} \\
Tangential Action &              & \bf 5125.6 {\scriptsize[$3423.8$, $7147.0$] }\\
\bottomrule
\end{tabular}
\captionof{table}{{\bf Effectiveness of tangential actions.} We execute actions tangent to \dracaena leaves in the Tiling baseline because they reveal more space on average compare to random actions.}
\tablelabel{tagent-vs-random}
\end{figure}
\renewcommand{\arraystretch}{1.0} \vspace{8pt}
We chose horizontal actions for the Tiling baseline of vines because they on average reveal the most amount of space. In order to come up with a similar Tiling baseline for \dracaena, we observe that leaves are pushed aside more easily when the grabber moves tangent to the leaves. We verify that tangent actions are better than random actions by comparing average space revealed upon execution of actions from the two methods. As shown in \tableref{tagent-vs-random}, tangential actions reveal more space than random actions, so we use them in the Tiling baseline to test the effectiveness of PPG w/ SRPNet against this strong baseline.
 
\clearpage

\section{Visualizations}

\begin{figure*}[h!]
    \centering
    \small
    \setlength{\tabcolsep}{0.1em} \begin{tabular}{cccccc}
        \toprule
        \rotatebox[origin=l]{0}{\footnotesize(a) RGB Image} & 
        \rotatebox[origin=l]{0}{\footnotesize(b) Height Image} &
        \rotatebox[origin=l]{0}{\footnotesize(c) Ground} &
        \rotatebox[origin=l]{0}{\footnotesize(d) \model} &
        \rotatebox[origin=l]{0}{\footnotesize(e) \model} &
        \rotatebox[origin=l]{0}{\footnotesize(f) \model} \\
        \rotatebox[origin=l]{0}{\footnotesize} & 
        \rotatebox[origin=l]{0}{\footnotesize} &
        \rotatebox[origin=l]{0}{\footnotesize Truth} &
        \rotatebox[origin=l]{0}{\footnotesize (No Image)} &
        \rotatebox[origin=l]{0}{\footnotesize (No Action)} &
        \rotatebox[origin=l]{0}{\footnotesize (Full)} \\
        \midrule
\includegraphics[width=0.16\textwidth]{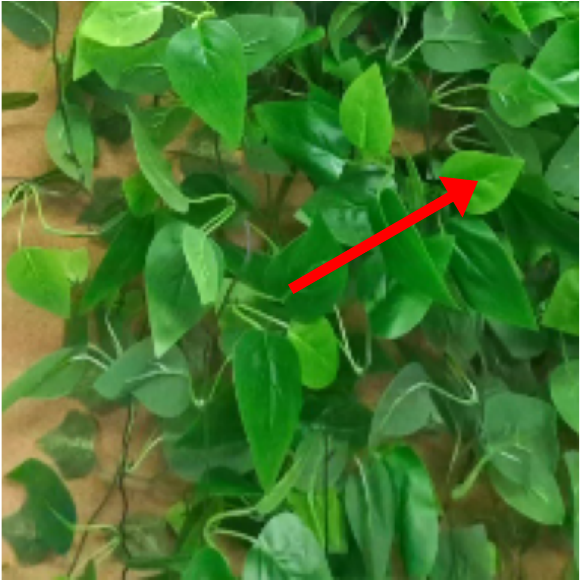} &
        \includegraphics[width=0.16\textwidth]{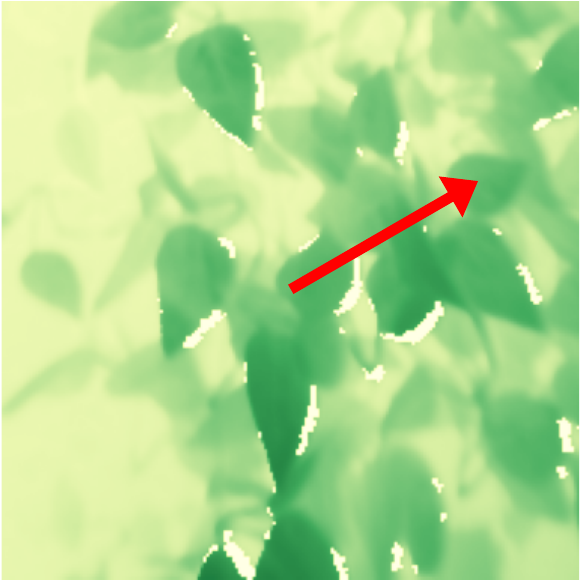}&
        \includegraphics[width=0.16\textwidth]{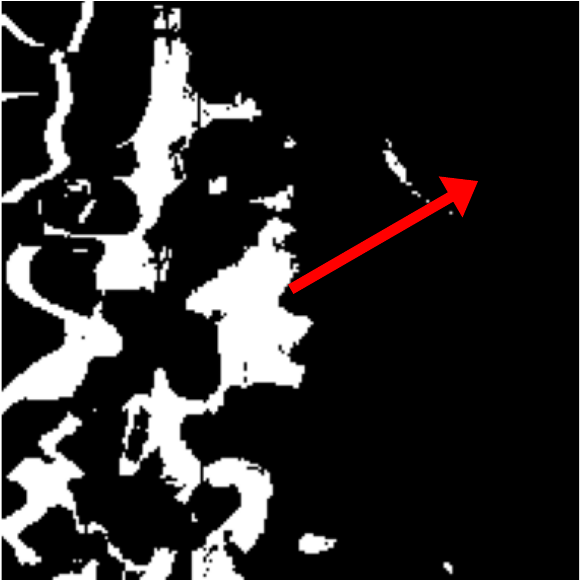}&
        \includegraphics[width=0.16\textwidth]{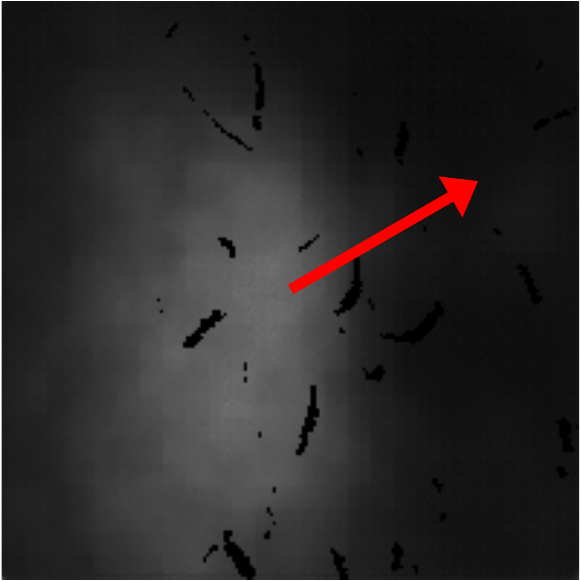}&
        \includegraphics[width=0.16\textwidth]{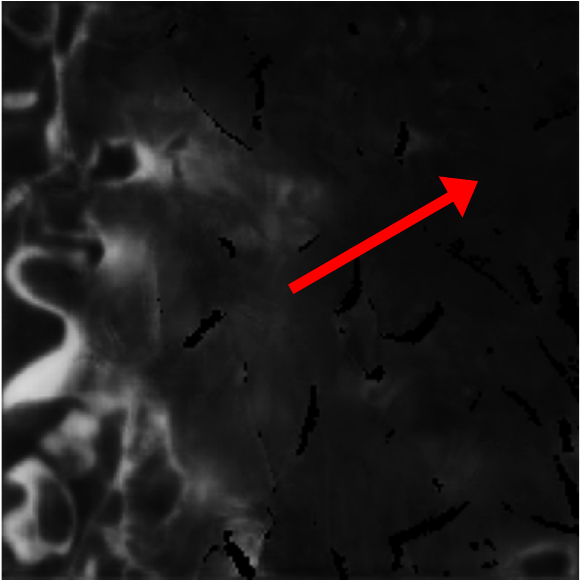}&
        \includegraphics[width=0.16\textwidth]{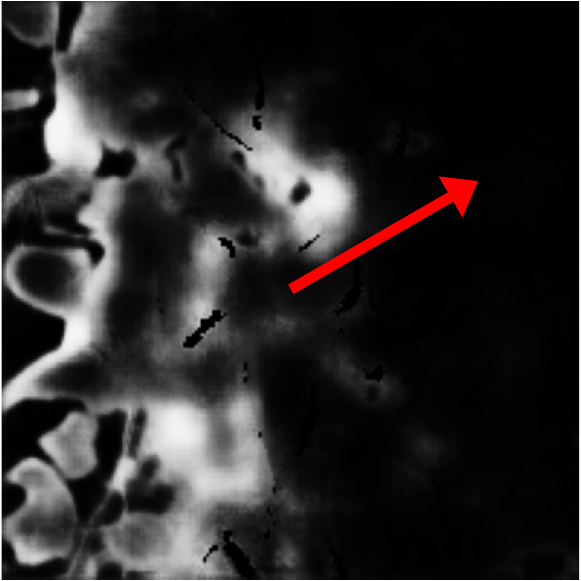} \\
        \includegraphics[width=0.16\textwidth]{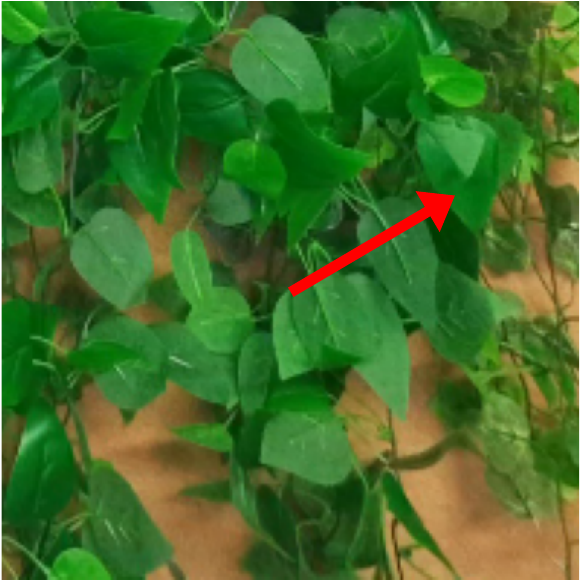} &
        \includegraphics[width=0.16\textwidth]{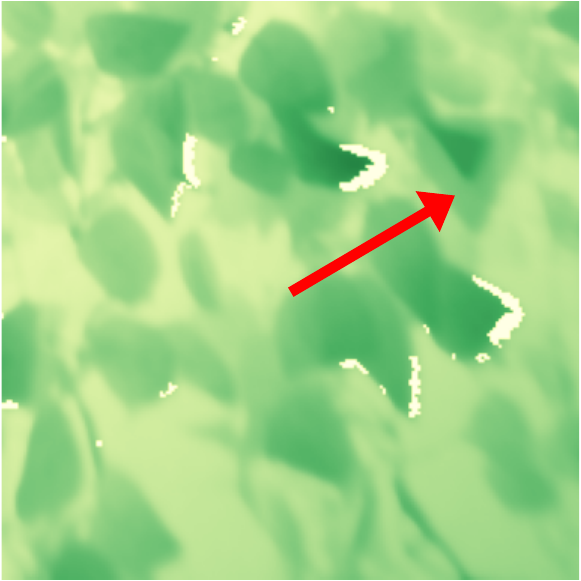}&
        \includegraphics[width=0.16\textwidth]{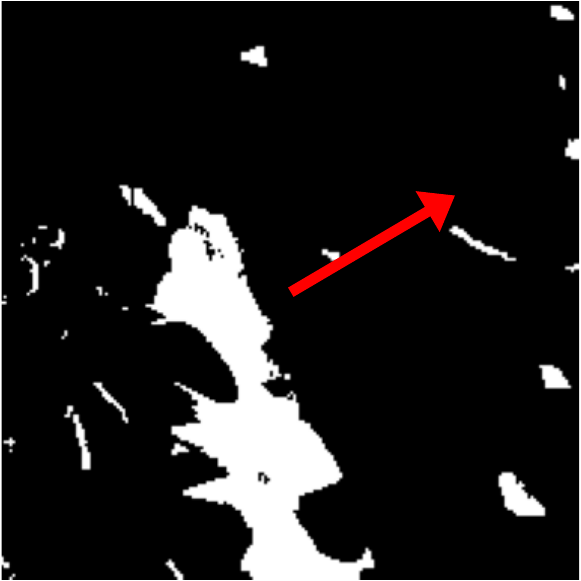}&
        \includegraphics[width=0.16\textwidth]{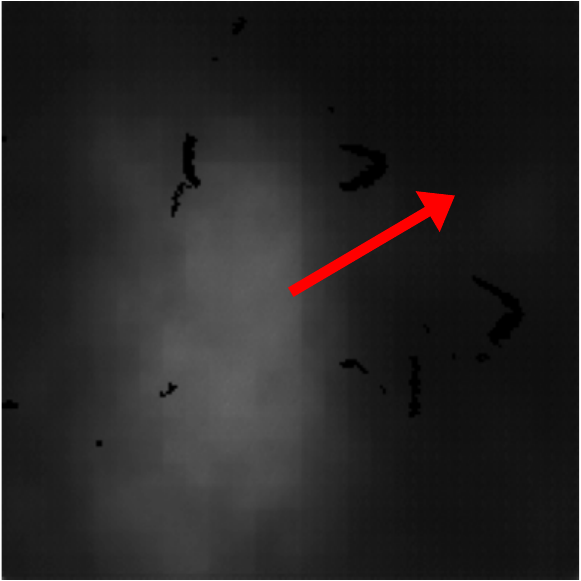}&
        \includegraphics[width=0.16\textwidth]{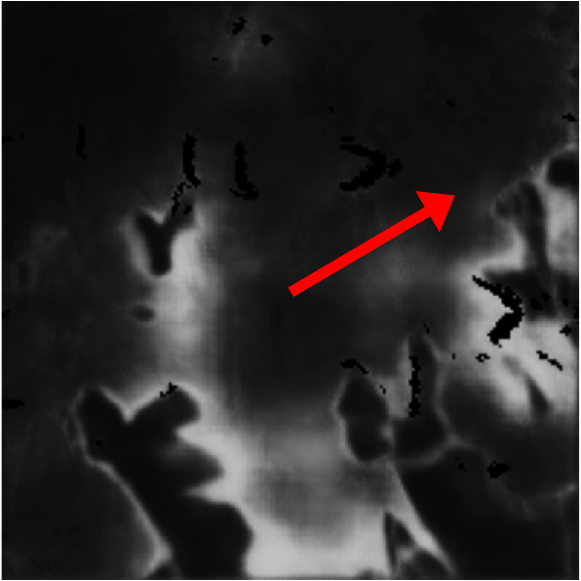}&
        \includegraphics[width=0.16\textwidth]{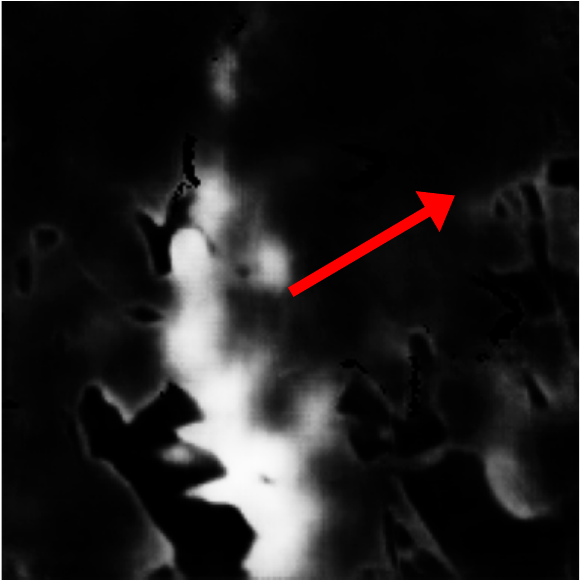} \\
        \includegraphics[width=0.16\textwidth]{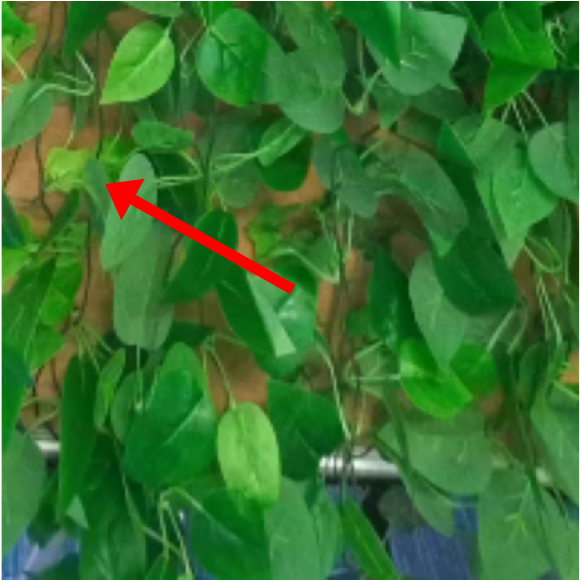} &
        \includegraphics[width=0.16\textwidth]{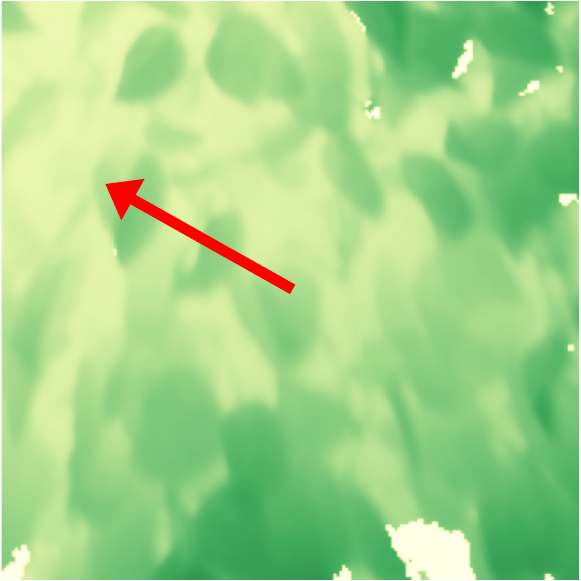}&
        \includegraphics[width=0.16\textwidth]{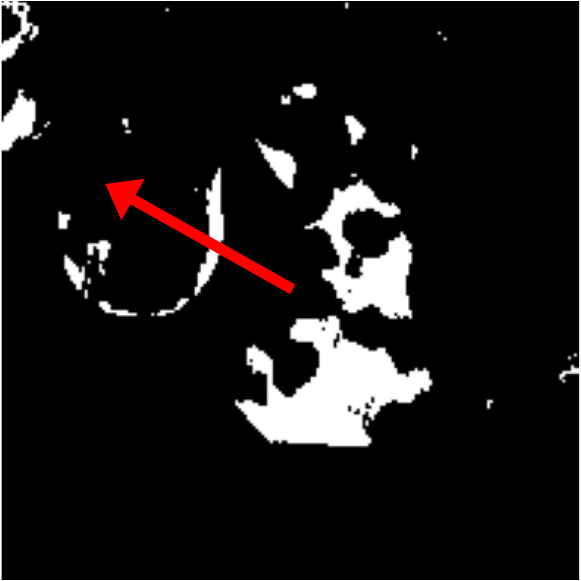}&
        \includegraphics[width=0.16\textwidth]{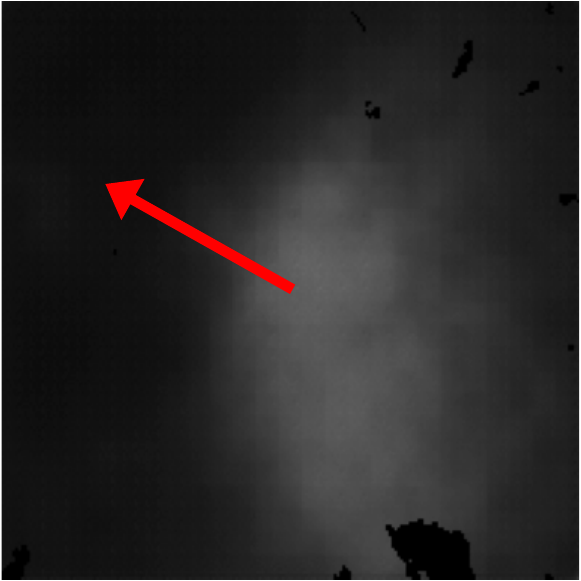}&
        \includegraphics[width=0.16\textwidth]{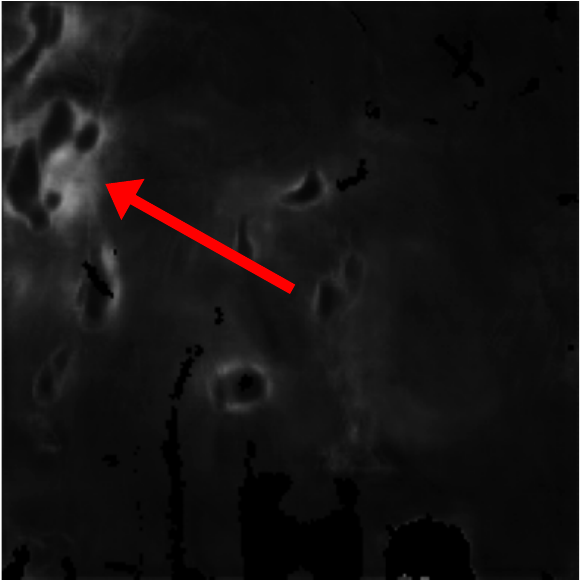}&
        \includegraphics[width=0.16\textwidth]{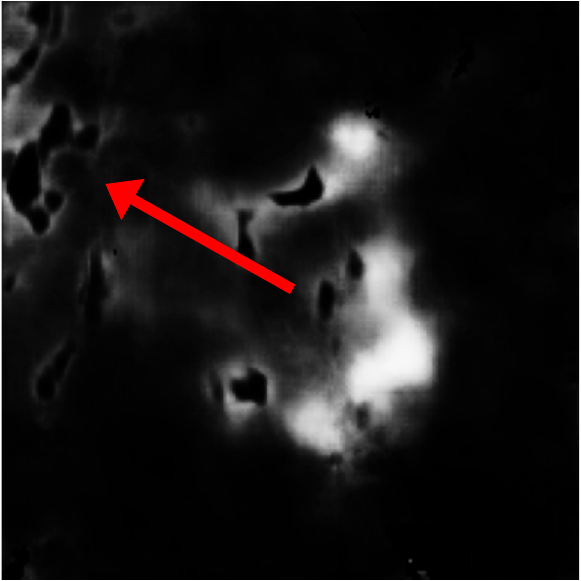} \\
        \midrule
        \includegraphics[width=0.16\textwidth]{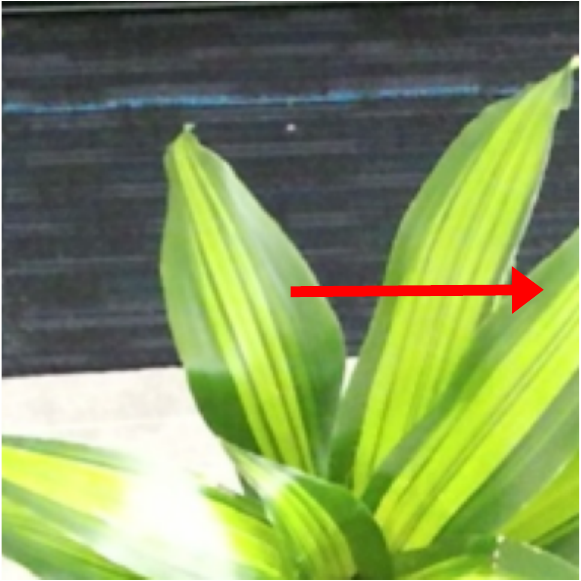} &
        \includegraphics[width=0.16\textwidth]{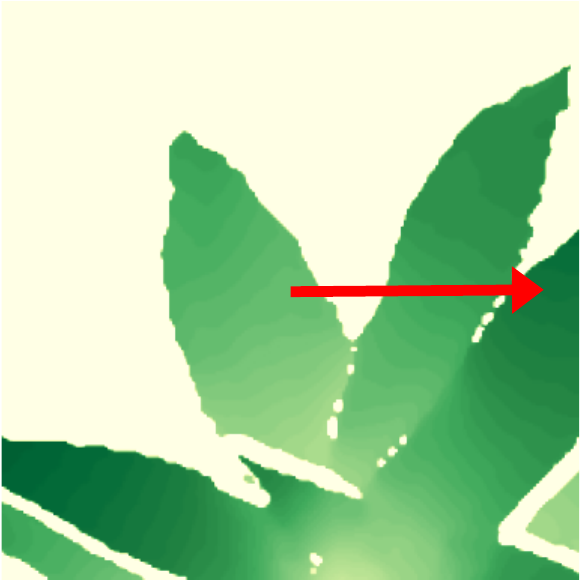} &
        \includegraphics[width=0.16\textwidth]{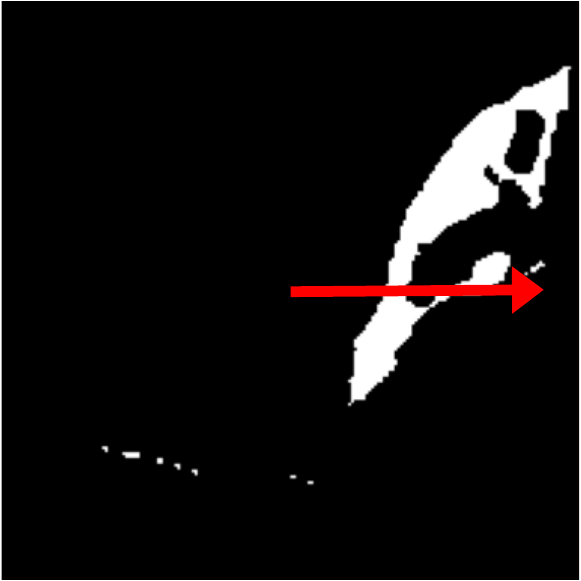} &
        \includegraphics[width=0.16\textwidth]{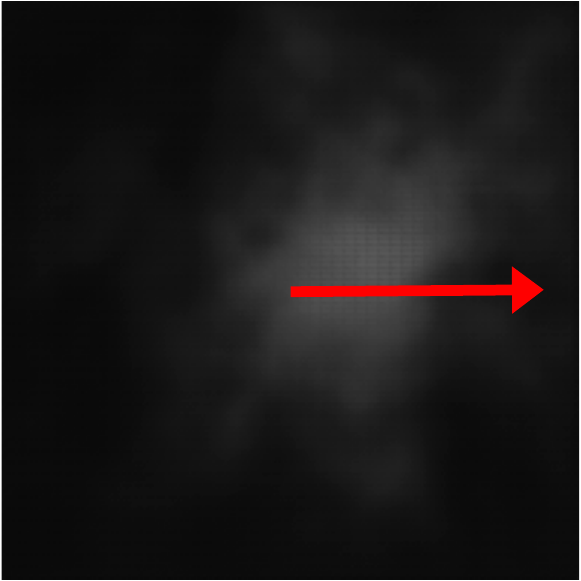} &
        \includegraphics[width=0.16\textwidth]{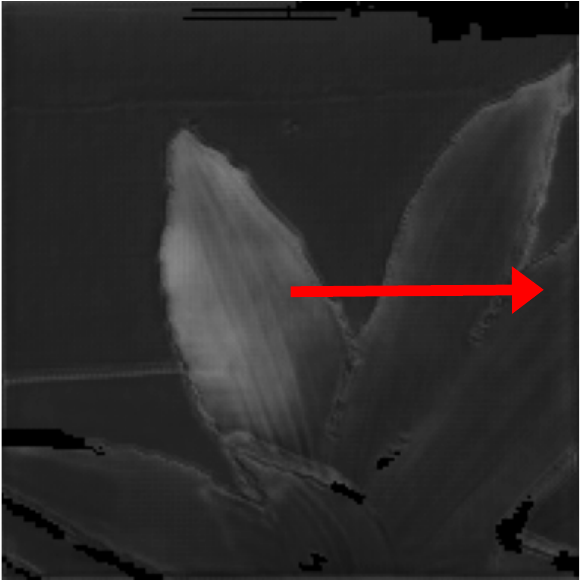} &
        \includegraphics[width=0.16\textwidth]{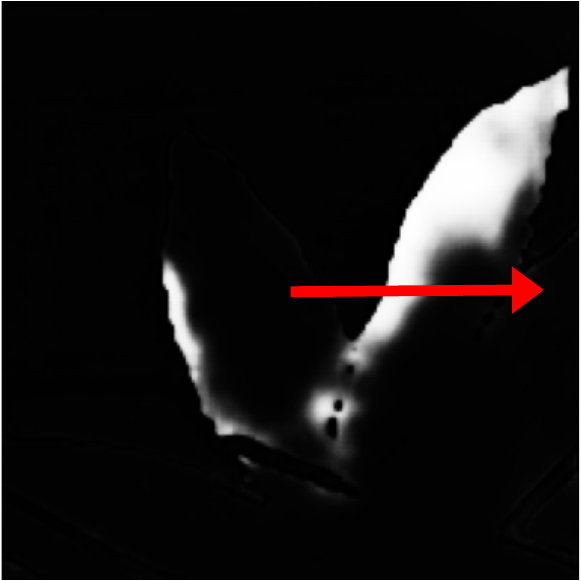} \\
        \includegraphics[width=0.16\textwidth]{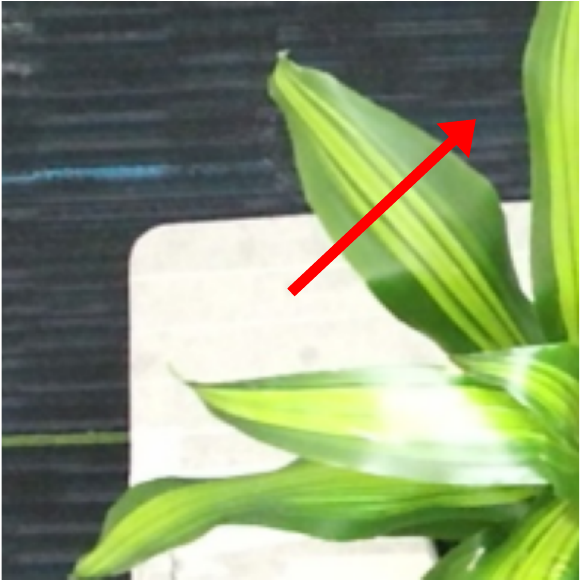} &
        \includegraphics[width=0.16\textwidth]{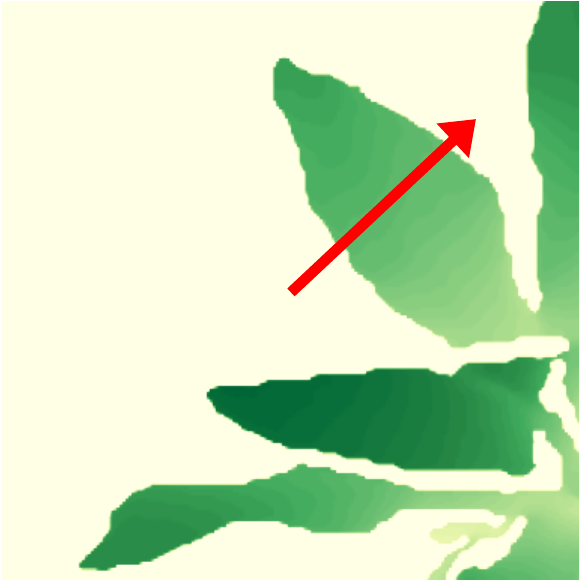} &
        \includegraphics[width=0.16\textwidth]{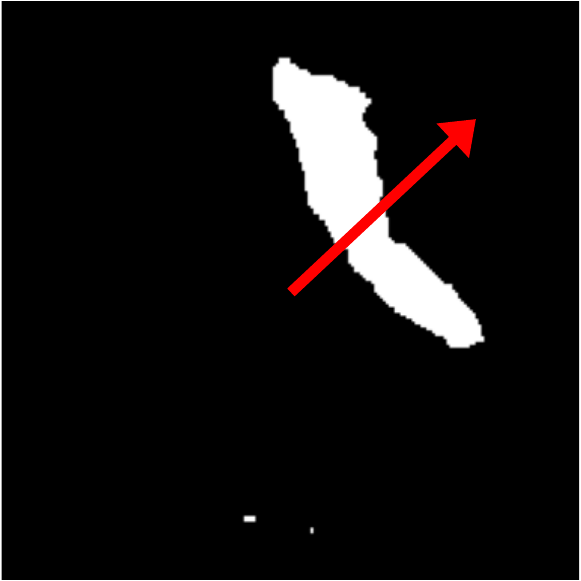} &
        \includegraphics[width=0.16\textwidth]{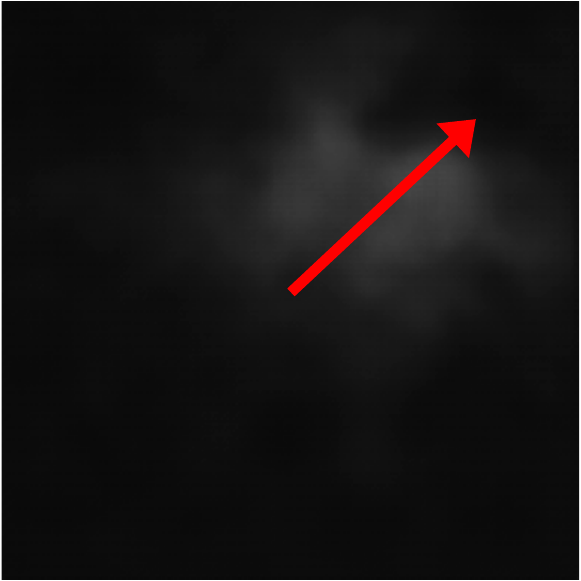} &
        \includegraphics[width=0.16\textwidth]{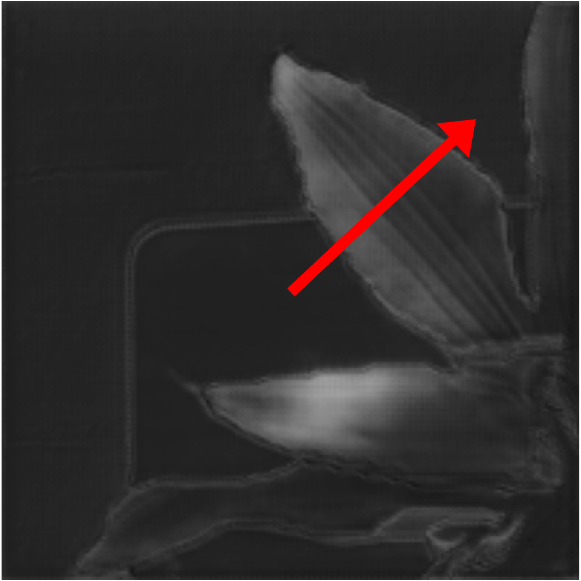} &
        \includegraphics[width=0.16\textwidth]{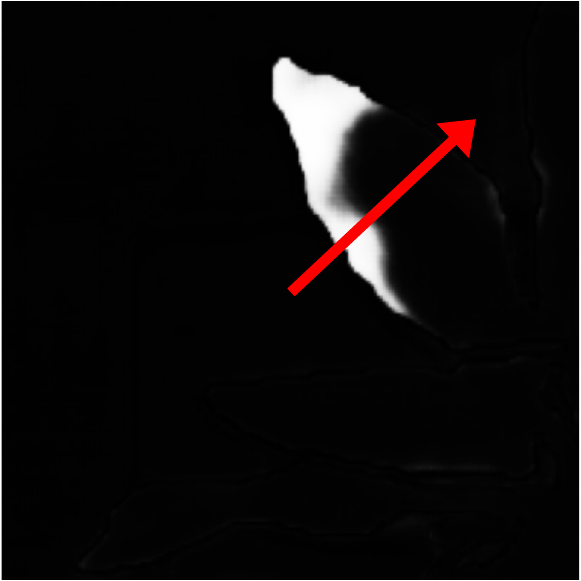} \\
        \includegraphics[width=0.16\textwidth]{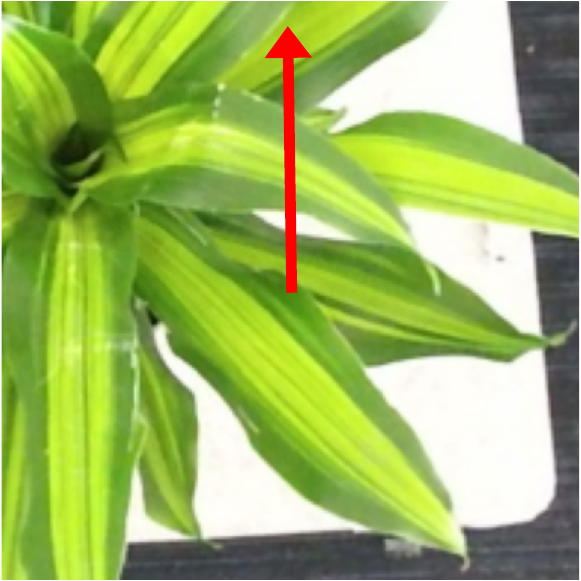} &
        \includegraphics[width=0.16\textwidth]{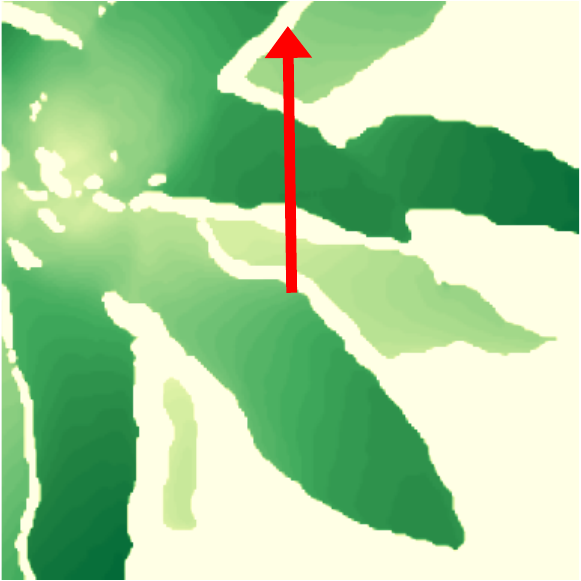} &
        \includegraphics[width=0.16\textwidth]{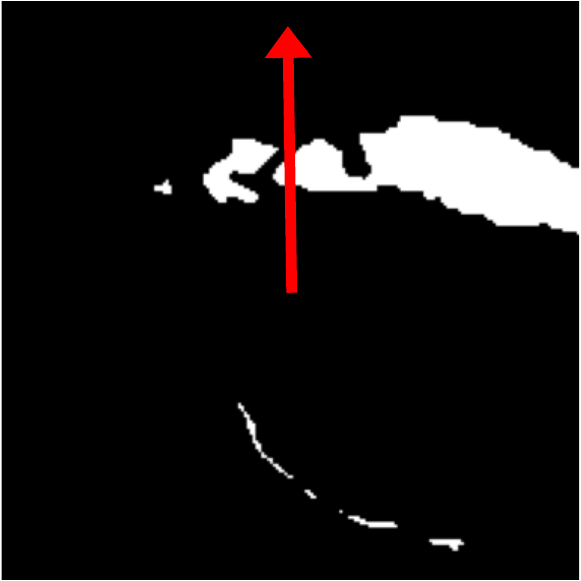} &
        \includegraphics[width=0.16\textwidth]{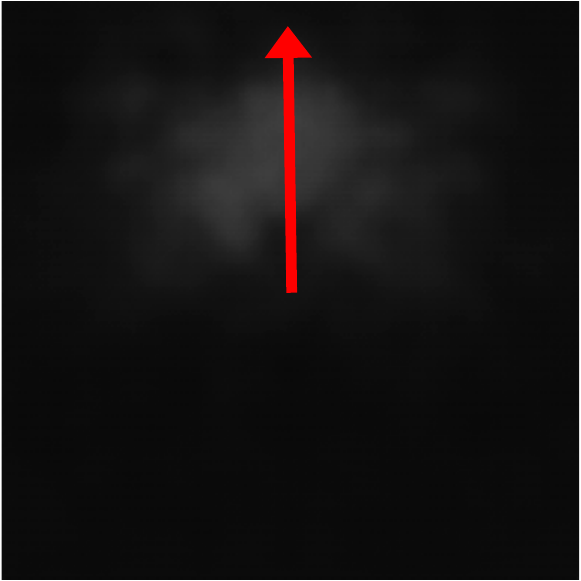} &
        \includegraphics[width=0.16\textwidth]{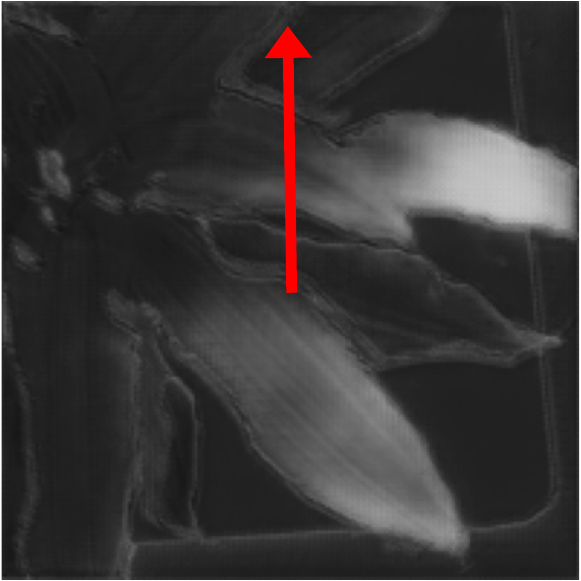} &
        \includegraphics[width=0.16\textwidth]{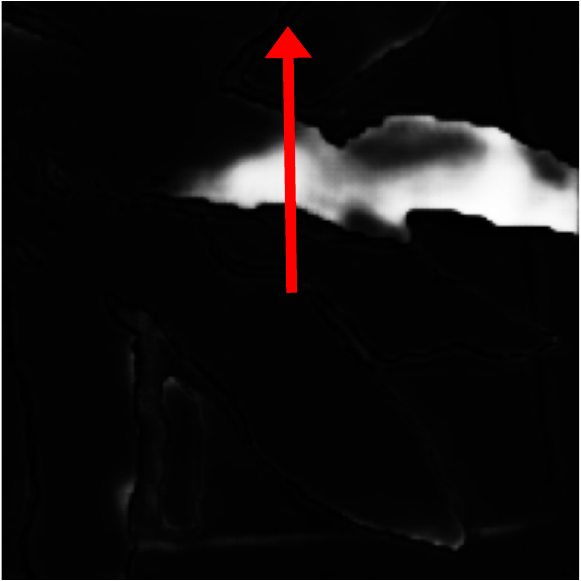} \\
        \includegraphics[width=0.16\textwidth]{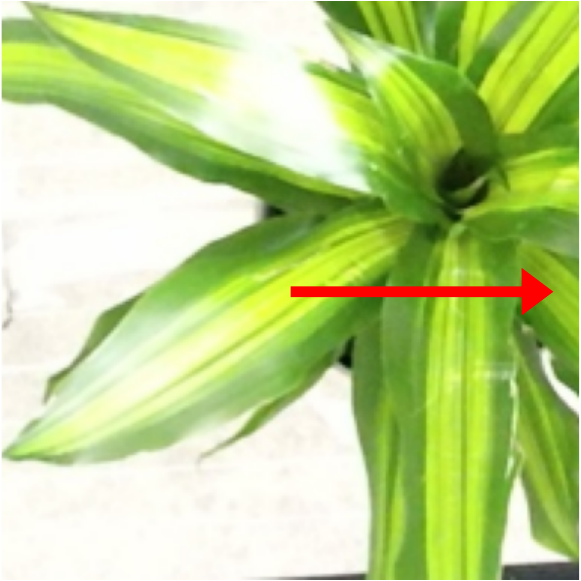} &
        \includegraphics[width=0.16\textwidth]{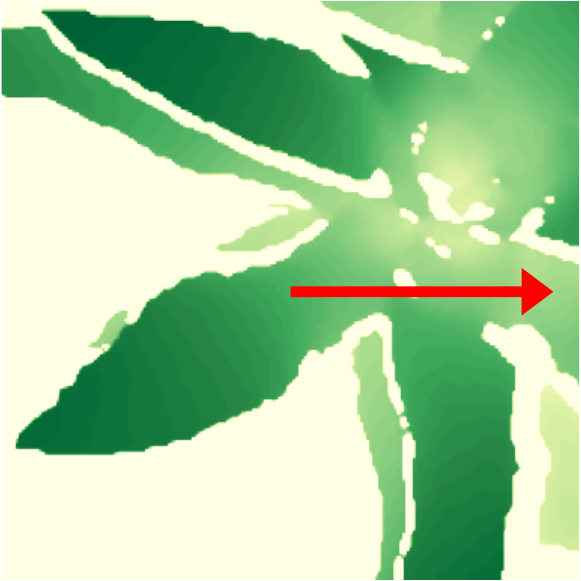} &
        \includegraphics[width=0.16\textwidth]{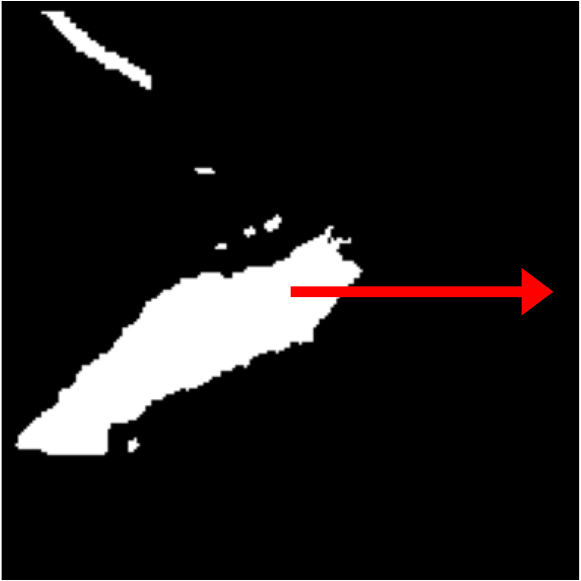} &
        \includegraphics[width=0.16\textwidth]{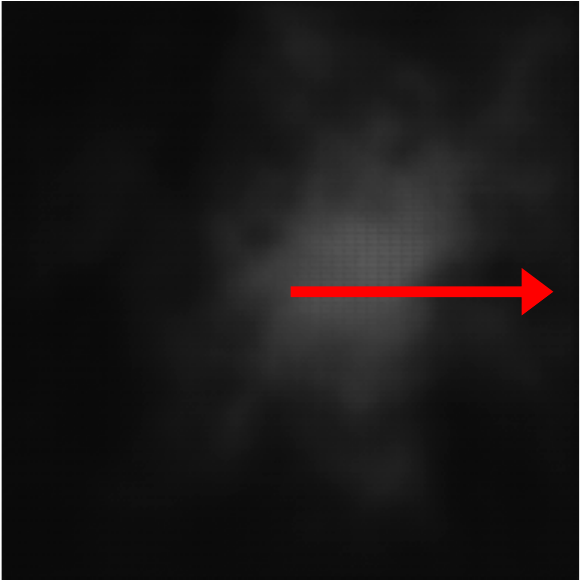} &
        \includegraphics[width=0.16\textwidth]{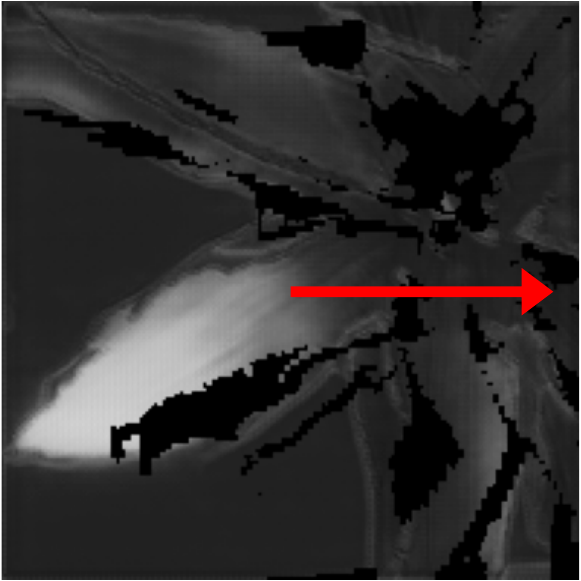} &
        \includegraphics[width=0.16\textwidth]{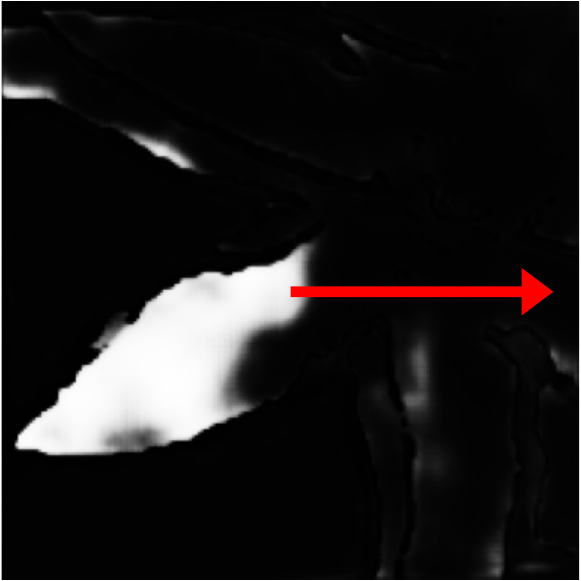} \\
        \bottomrule
    \end{tabular}
\caption{{\bf Visualizations of output from our proposed \model.} We show 
examples from the test set. The white regions in ground truth images represent
space revealed by actions drawn as red arrows. Column (d) shows prediction from \model without image input (\ie no \rgb, no height), column (e) shows prediction from \model without action input, and column (f) shows predictions from \model. The brighter the region, the higher the predicted probability of revealing space. Ground truth revealed space indicates the complexity of the task and suggests why the hand-crafted dynamics model (shown in \figref{spaghetti-model}) performs poorly at this task. SRPNet is able to effectively use the visual information to make good predictions.}
\figlabel{vis-forward-model}
\end{figure*}

\setlength{\fboxsep}{0pt}
\begin{figure*}
\centering
    \small
    \setlength{\tabcolsep}{0.1em} \begin{tabular}{cccccc}
        \toprule
         & $t=0$ & $t=1$ & $t=2$ & $t=3$ & $t=4$ \\
        \midrule
\rotatebox[origin=l]{90}{\footnotesize View before Interaction} &
        \includegraphics[width=0.19\linewidth]{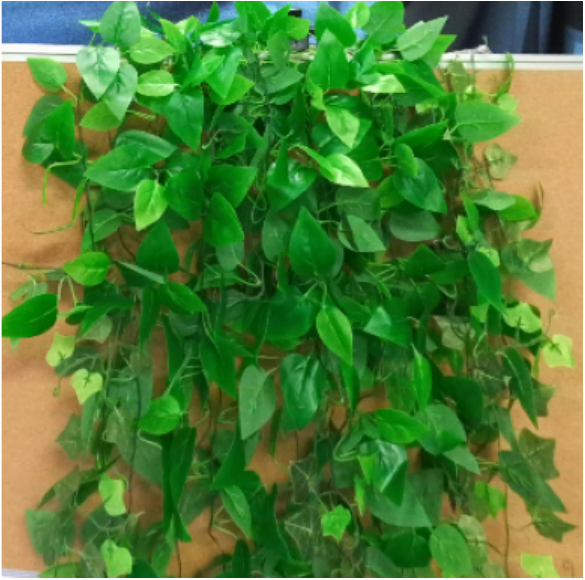} &
        \includegraphics[width=0.19\linewidth]{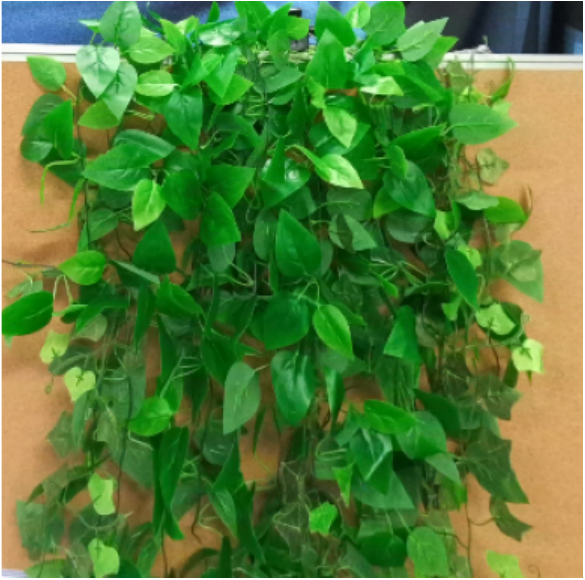} & 
        \includegraphics[width=0.19\linewidth]{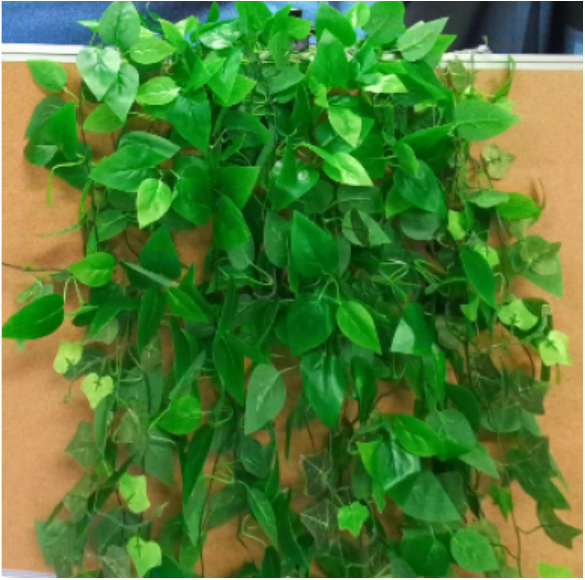} &
        \includegraphics[width=0.19\linewidth]{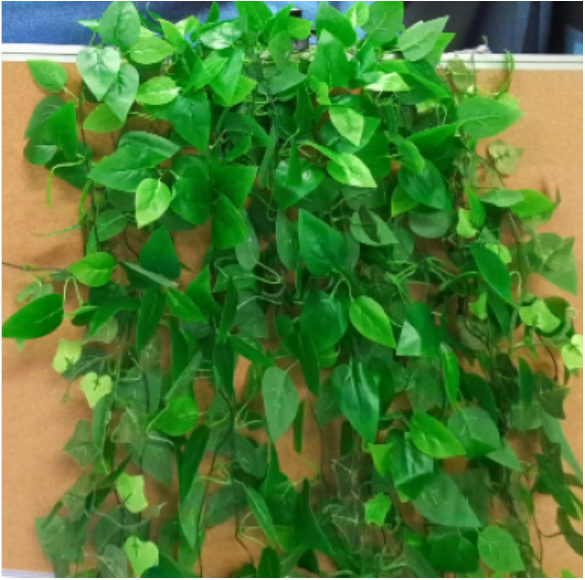} &
        \includegraphics[width=0.19\linewidth]{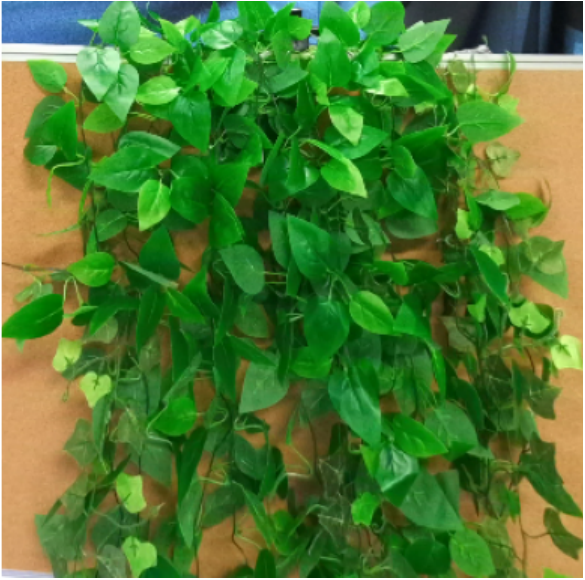}\\
\rotatebox[origin=l]{90}{\footnotesize Executed Action} &
        \includegraphics[width=0.19\linewidth]{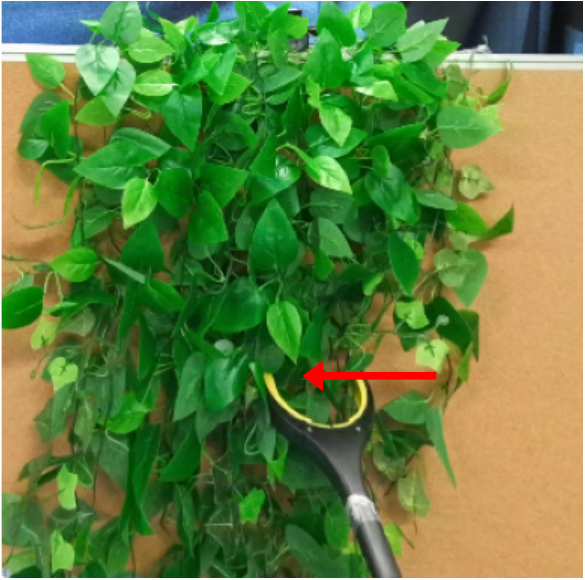} &
        \includegraphics[width=0.19\linewidth]{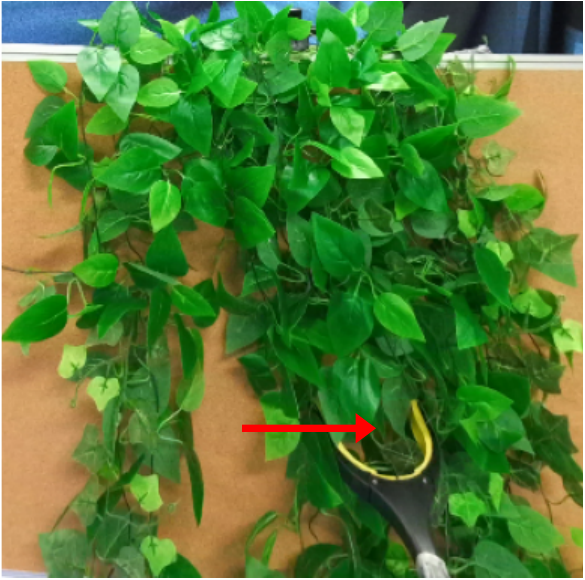} & 
        \includegraphics[width=0.19\linewidth]{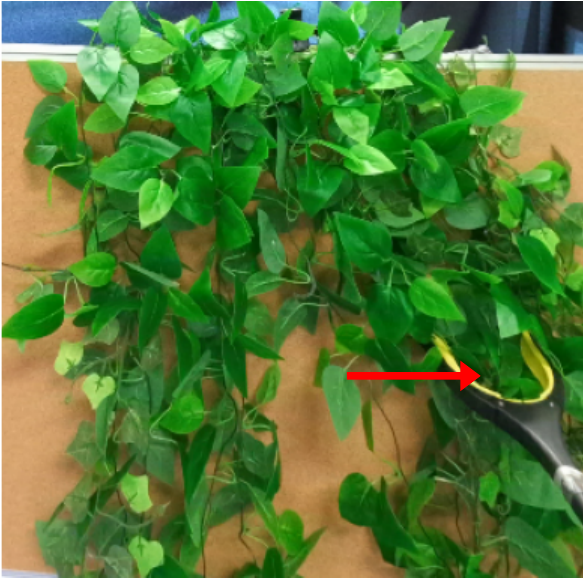} &
        \includegraphics[width=0.19\linewidth]{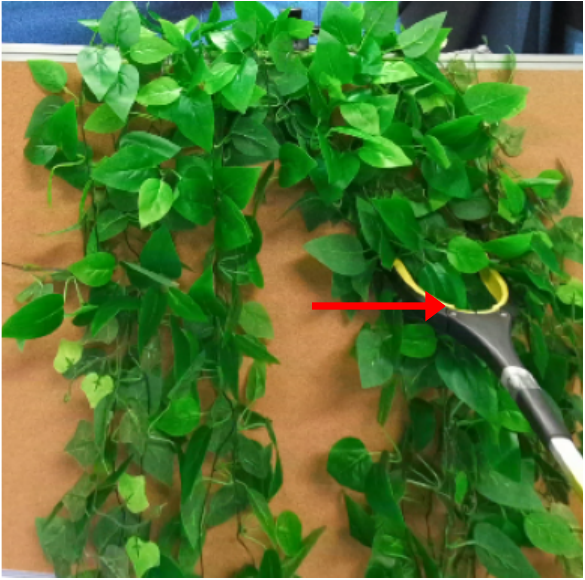} & 
        \includegraphics[width=0.19\linewidth]{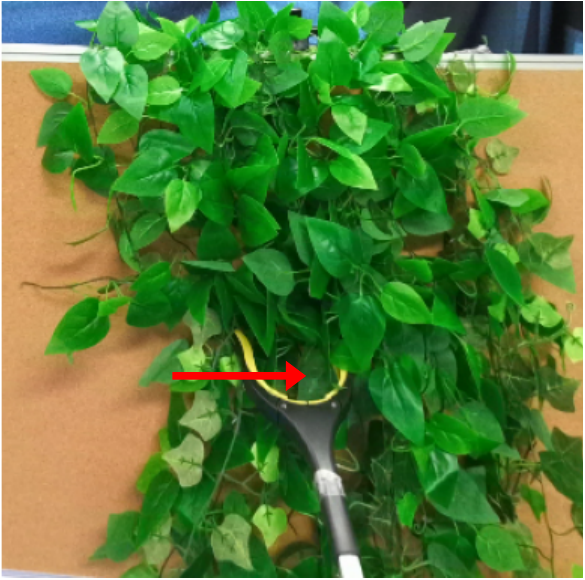} \\
\rotatebox[origin=l]{90}{\footnotesize Space Revealed ($C_{t+1})$} &
        \includegraphics[width=0.19\linewidth,fbox]{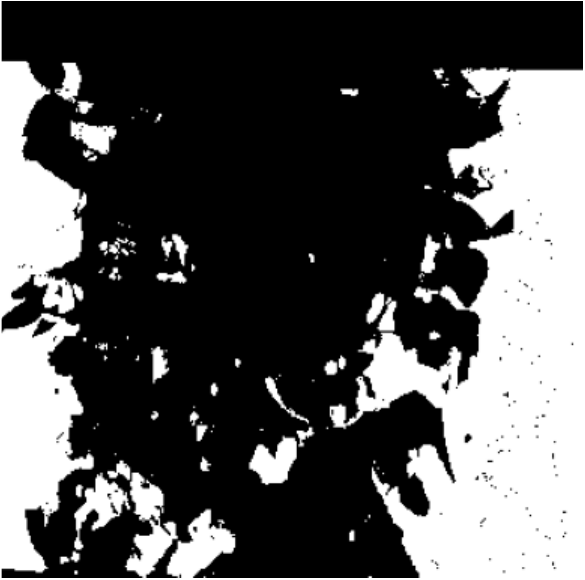} & 
        \includegraphics[width=0.19\linewidth,fbox]{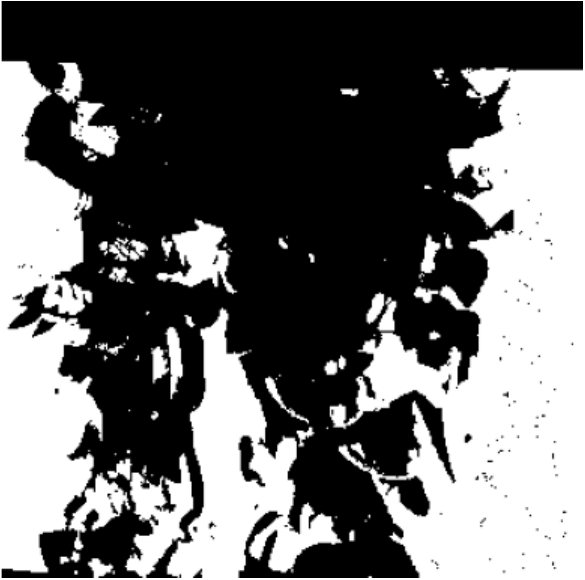} & 
        \includegraphics[width=0.19\linewidth,fbox]{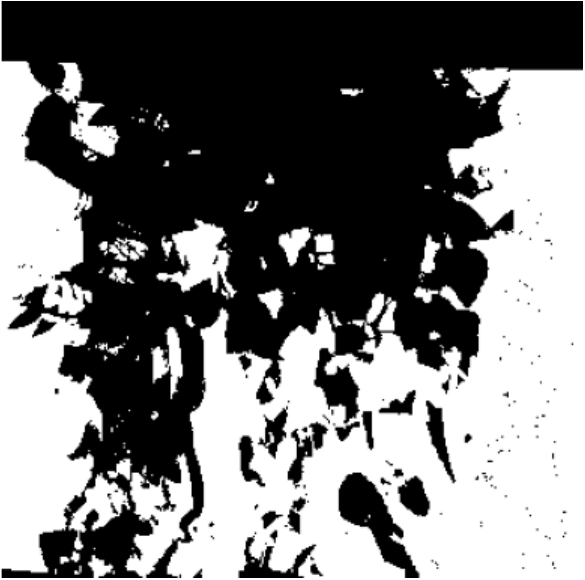} & 
        \includegraphics[width=0.19\linewidth,fbox]{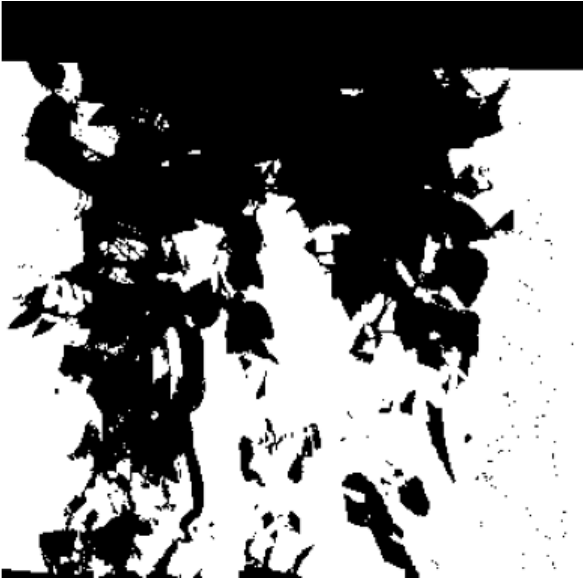} & 
        \includegraphics[width=0.19\linewidth,fbox]{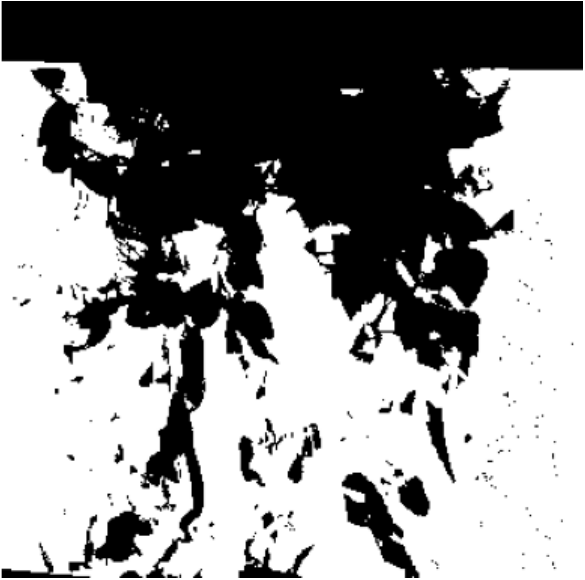} \\
        \midrule
        \rotatebox[origin=l]{90}{\footnotesize View before Interaction} &
        \includegraphics[width=0.19\linewidth]{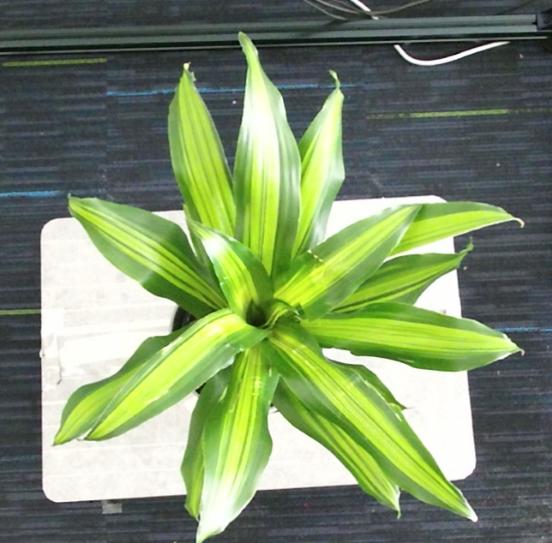} &
        \includegraphics[width=0.19\linewidth]{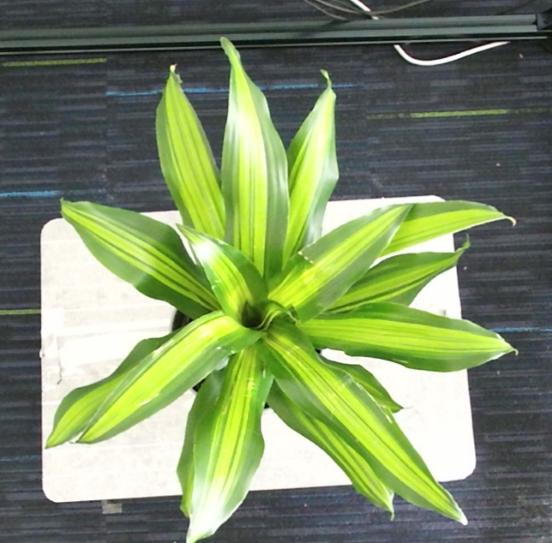} & 
        \includegraphics[width=0.19\linewidth]{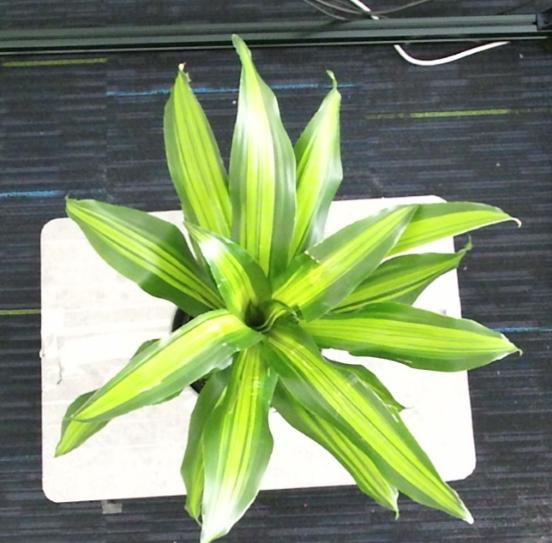} &
        \includegraphics[width=0.19\linewidth]{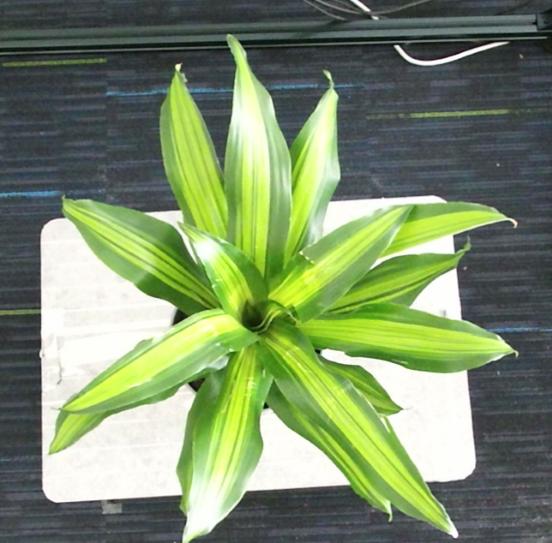} &
        \includegraphics[width=0.19\linewidth]{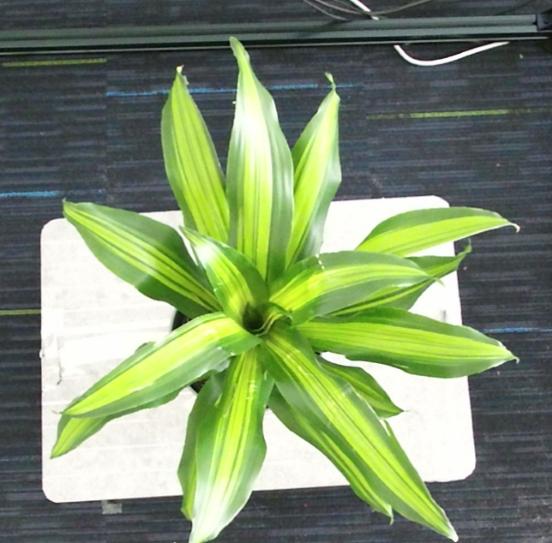}\\
        \rotatebox[origin=l]{90}{\footnotesize Executed Action} &
        \includegraphics[width=0.19\linewidth]{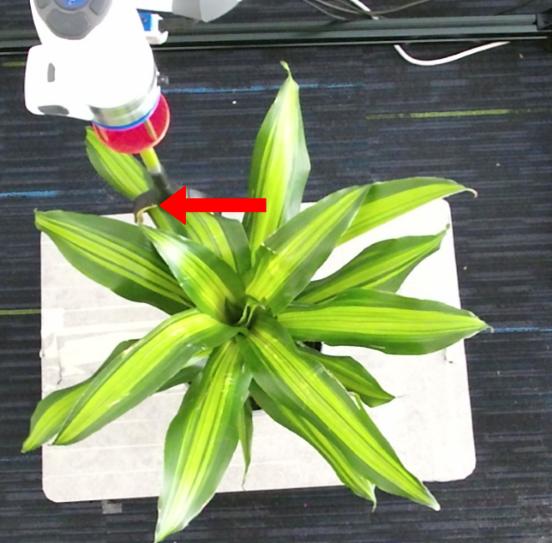} &
        \includegraphics[width=0.19\linewidth]{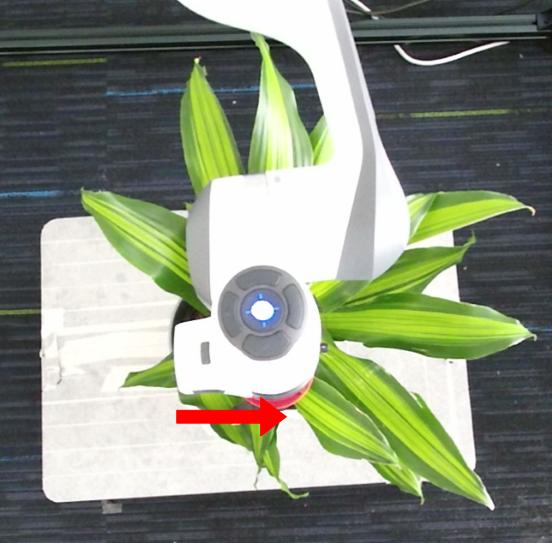} & 
        \includegraphics[width=0.19\linewidth]{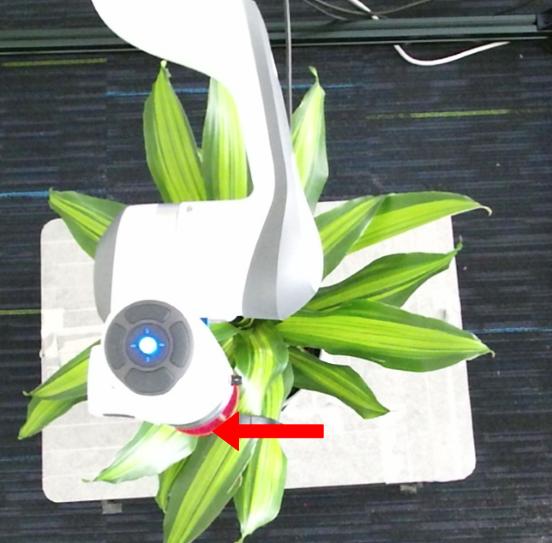} &
        \includegraphics[width=0.19\linewidth]{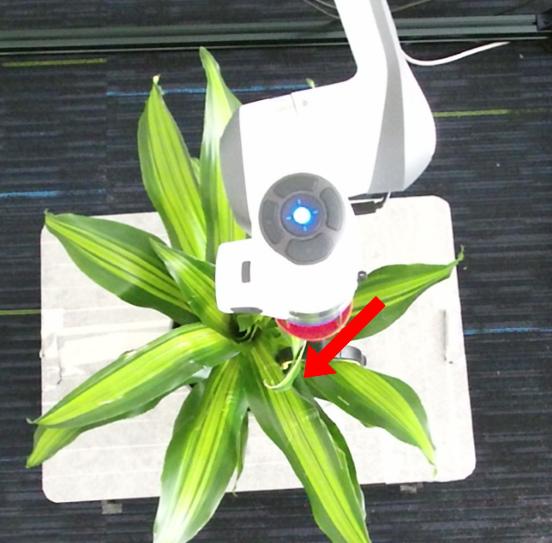} &
        \includegraphics[width=0.19\linewidth]{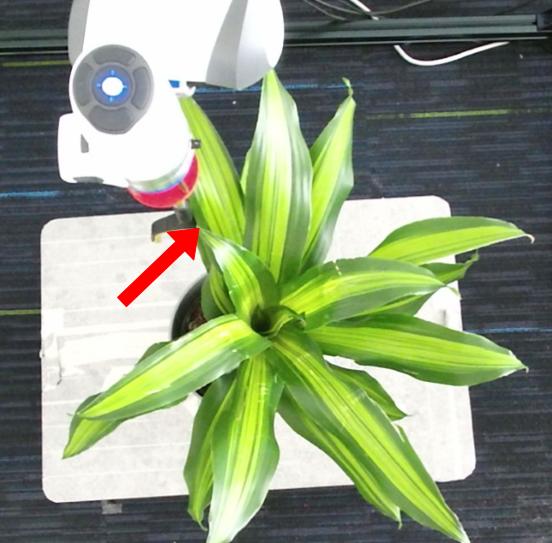}\\
        \rotatebox[origin=l]{90}{\footnotesize Space Revealed ($C_{t+1})$} &
        \includegraphics[width=0.19\linewidth,fbox]{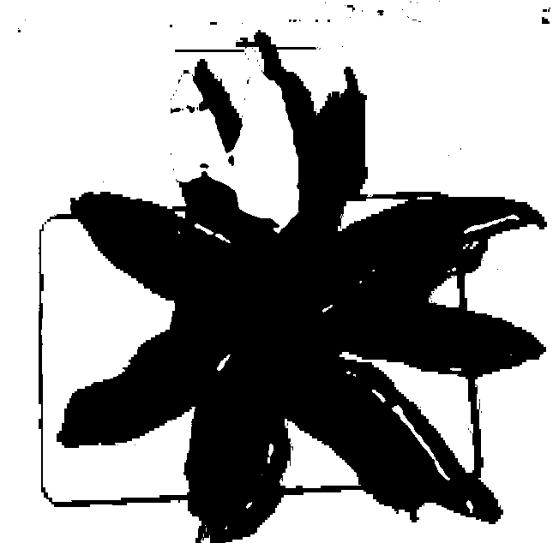} & 
        \includegraphics[width=0.19\linewidth,fbox]{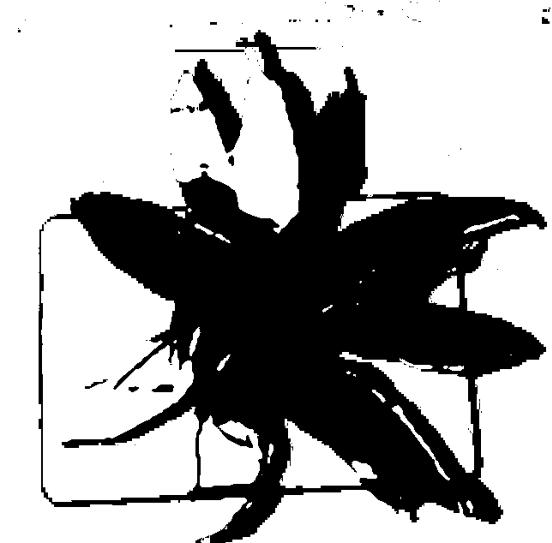} & 
        \includegraphics[width=0.19\linewidth,fbox]{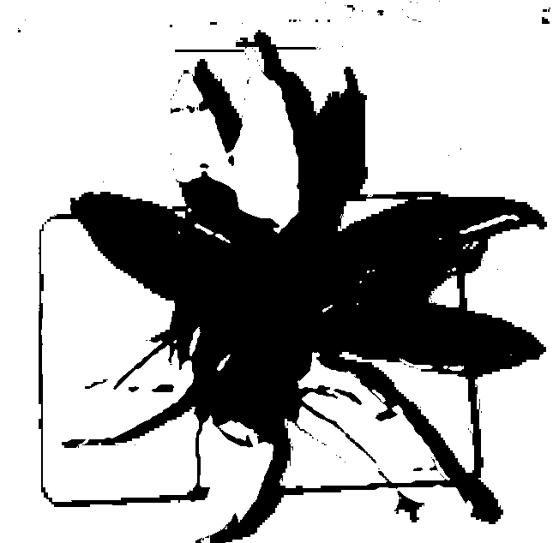} & 
        \includegraphics[width=0.19\linewidth,fbox]{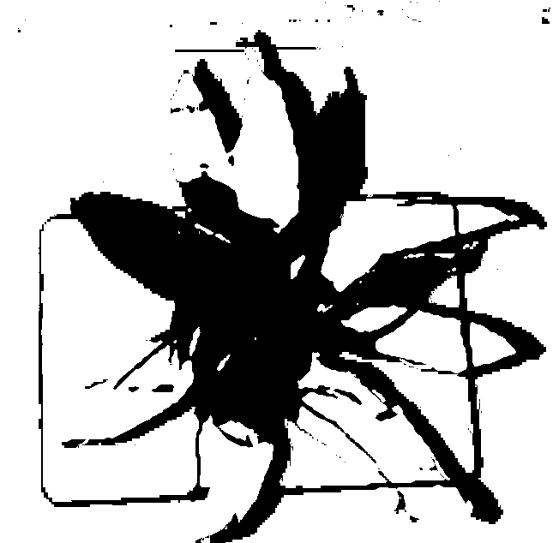} & 
        \includegraphics[width=0.19\linewidth,fbox]{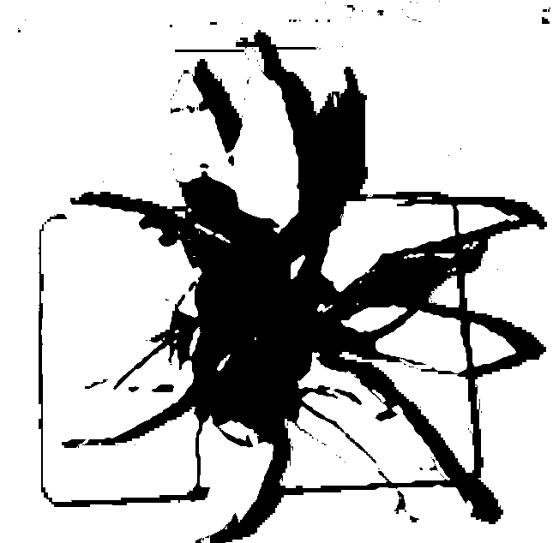} \\
        \bottomrule
    \end{tabular}
\caption{{\bf First five time steps of a sample execution from our method.} Top row shows the \rgb image before interaction, middle row shows the push action executed, and the bottom row shows the cumulative space revealed so far. Our model picks actions that are effective at revealing space.}
\figlabel{vis-execution}
\end{figure*}

\end{document}